\documentclass{article}

\usepackage{PRIMEarxiv}

\usepackage[utf8]{inputenc} 
\usepackage[T1]{fontenc}    
\usepackage{hyperref}       
\usepackage{url}            
\usepackage{booktabs}       
\usepackage{amsfonts}       
\usepackage{nicefrac}       
\usepackage{microtype}      
\usepackage{lipsum}
\usepackage{fancyhdr}       
\usepackage{graphicx}       
\graphicspath{{media/}}     
\usepackage[normalem]{ulem}
\useunder{\uline}{\ul}{}
\usepackage[table]{xcolor}
\usepackage{amsmath}
\usepackage{amsfonts}
\usepackage{todonotes}
\usepackage{adjustbox}
\usepackage{subcaption}
\newcommand{\pulrad}[1]{\raisebox{1.5ex}[0pt]{#1}}
\usepackage[square,sort,comma,numbers]{natbib}

\title{Exploring Multiple Strategies to Improve Multilingual Coreference Resolution in CorefUD
}

\author{
  Ond\v{r}ej Pra\v{z}\'ak \\
  New Technologies for the Information Society,\\ Faculty of Applied Sciences, \\
  University of West Bohemia \\
  Pilsen\\
  \texttt{ondfa@ntis.zcu.cz} \\
  \And
  Miloslav Konop\'ik \\
  Department of Computer Science and Engineering,\\ Faculty of Applied Sciences, \\
  University of West Bohemia \\
  Pilsen\\
  \texttt{konopik@kiv.zcu.cz} \\
  \And
  Pavel Kr\'al \\
  Department of Computer Science and Engineering,\\ Faculty of Applied Sciences, \\
  University of West Bohemia \\
  Pilsen\\
  \texttt{pkral@kiv.zcu.cz} \\
}

\begin{document}
\maketitle

\begin{abstract}
Coreference resolution, the task of identifying expressions in text that refer to the same entity, is a critical component in various natural language processing applications. This paper presents a novel end-to-end neural coreference resolution system, utilizing the CorefUD 1.1 dataset, which spans 17 datasets across 12 languages. The proposed model is based on the standard end-to-end neural coreference resolution system. We first establish baseline models, including monolingual and cross-lingual variations, and then propose several extensions to enhance performance across diverse linguistic contexts. These extensions include cross-lingual training, incorporation of syntactic information, a Span2Head model for optimized headword prediction, and advanced singleton modeling. We also experiment with headword span representation and long-documents modeling through overlapping segments. The proposed extensions, particularly the heads-only approach, singleton modeling,  and long document prediction, significantly improve performance across most datasets. We also perform zero-shot cross-lingual experiments, highlighting the potential and limitations of cross-lingual transfer in coreference resolution. Our findings contribute to the development of robust and scalable coreference systems for multilingual coreference resolution. Finally, we evaluate our model on the CorefUD 1.1 test set and surpass the best model from the CRAC 2023 shared task of comparable size by a large margin.
\end{abstract}

\keywords{coreference resolution \and cross-lingual model \and Transformers \and end-to-end model}

\section{Introduction}

Coreference resolution is the task of identifying which expressions in a text refer to the same entity in the real world. The task can be divided into two subtasks. Identify entity mentions and group them according to the entities they refer to. Anaphora resolution is closely related to coreference resolution. Anaphora is a contextual reference to an expression in the previous context. Most of the anaphoric relations are also coreferential, but it does not necessarily be so. For example, in the sentence: 

\textit{\textbf{This apple} is mine. Please, take another \textbf{one}},

\noindent the two mentions are anaphoric, but they are not coreferential (they refer to different apples). Some datasets contain only anaphora, some only coreferences, and others contain annotations of both. Anaphoric relations are based more on the syntax of a sentence, whereas coreference relations are more about semantics.
Coreferences are usually annotated in terms of coreference clusters or in terms of coreference chains. Coreference clusters group together all the mentions of the same entity, no matter their order. In the coreference chain, each mention is aligned with its direct antecedent (closest previous mention of the same entity).
Although significant advancements have been made in coreference resolution for English, developing robust and effective models for multiple languages remains a challenging task due to linguistic diversity and the availability of resources.


Historically, coreference resolution was a standard preprocessing step in various natural language processing (NLP) tasks, such as machine translation, summarization, and information extraction. Although recent large language models have achieved state-of-the-art results in coreference resolution, they are expensive to train and deploy, and traditional (discriminative) approaches remain competitive. Expressing this task in natural language is challenging, and to the best of our knowledge, there have been no successful attempts to utilize large chatbots (like ChatGPT-4) to achieve superior results.




Coreference resolution becomes particularly challenging in low-resource languages. One strategy to address this challenge is to train a multilingual model on datasets from multiple languages, thereby transferring knowledge from resource-rich languages to those with fewer resources. However, a significant challenge with this approach lies in the differences in annotations across available corpora. The CorefUD initiative \cite{corefud2022lrec} tries to harmonize the datasets and create one annotation scheme for coreference in multiple languages similarly to Universal Dependencies \cite{nivre-etal-2020-universal} for syntactic annotations. 

This paper proposes a novel approach to multilingual coreference resolution. We use the CorefUD 1.1 dataset \cite{corefud2022lrec}. The CorefUD 1.1 is a harmonized corpus with 17 different datasets spanning 12 languages. The dataset is intended to extend Universal Dependencies \cite{nivre-etal-2020-universal} with coreference annotations. Therefore, all datasets within CorefUD are treebanks. Coreference annotations are built upon these dependencies, meaning that mentions are represented as subtrees in the dependency tree and can be captured by their heads. Moreover, the evaluation metric used with this dataset on CRAC 2023 Multilingual Coreference Resolution Shared Task uses head-match evaluation. This means that the predicted mention is considered the same as the gold mention if they share the same head. Notable differences exist between the datasets, with one of the most prominent being the presence of singletons. Singletons are clusters that contain only one mention and do not participate in any coreference relation, yet they are still annotated as mentions. For further details on the dataset, refer to \cite{corefud2022lrec}.

Building upon the baseline model proposed by \cite{xu-choi-2020-revealing}, we introduce several novel extensions aimed at enhancing the performance of coreference resolution across multiple languages and datasets within the CorefUD collection. Our primary goal is to develop a universal model capable of handling the diverse and complex nature of these datasets effectively.

First, recognizing the challenge posed by small dataset sizes in the CorefUD collection, we propose a cross-lingual training approach. By pretraining our model on a concatenated dataset that includes all available training data across languages, we aim to improve the model's ability to generalize across languages. This approach is particularly beneficial for low-resource languages, where training large models from scratch is impractical due to the limited data available.

Next, we incorporate syntactic information into our model to leverage the dependency structures inherent in the CorefUD datasets. This extension adds syntax to the token representations by encoding their paths to the ROOT in the dependency tree, thereby enriching the model’s understanding of syntactic relationships, which are crucial for accurate mention detection and coreference resolution.

In response to the evaluation metrics used in the CRAC 2022 Shared Task, we also evaluate the Span2head model, which simplifies span-based mention representation to headword prediction. This adjustment aligns the model's output more closely with the evaluation criteria, where we consider two mentions the same if they share the same head.

Additionally, we explore the potential of head representations as a simplified approach to coreference resolution. Similarly to word-level coreference \cite{dobrovolskii2021word}, by modeling mentions solely based on their syntactic heads, we reduce the computational complexity from quadratic to linear, making the model more efficient and less prone to errors, particularly in the case of long and complex mentions.

We further address the issue of singleton—mentions by introducing mechanisms to incorporate these into the training process. By modeling singletons explicitly, we ensure that valuable training data from singleton-rich datasets is not discarded, thus improving the model’s robustness and accuracy across different datasets.

Finally, to overcome the limitations of short sequence lengths in XLM-R, we propose an approach that utilizes overlapping segments with a cluster merging algorithm. This method ensures that coreference chains spanning multiple segments are correctly identified and merged, even when document segmentation is necessary due to memory constraints.

We experimentally evaluate all proposed extensions to find the best configuration for a joined model for all the datasets in CorefUD. Through these extensions, our model aims to advance the state of multilingual coreference resolution by addressing the specific challenges posed by the diverse datasets in the CorefUD collection, ultimately contributing to more accurate and generalizable coreference systems.

The main contributions of the paper are the following:

\begin{itemize}
    \item We propose several extensions of a standard model that lead to significant improvements. The most prominent extensions include singletons modeling, headword mention representation, and segments overlapping for long coreference prediction.
    \item We provide a detailed evaluation of all proposed extensions.
    \item We perform the analysis of knowledge transferability between the datasets with joined training and cross-lingual zero-shot experiments.
    \item We demonstrated that the proposed model significantly surpasses the state of the art with a comparable model size.
    \item We trained a single model that has a great performance on all the datasets in the CorefUD collection.
    \item Our source codes are publicly available on GitHub for subsequent research: \url{https://github.com/ondfa/coref-multiling}.
\end{itemize}

\section{Related Work}



\subsection{End-to-end Neural Coreference Resolution Models}

The first end-to-end neural coreference resolution system was introduced by \cite{lee-etal-2017-end}, and many subsequent neural coreference resolution systems are based on their model. 

The model estimates the probability $P(y_i|D)$ of a mention $i$ corefering with the antecedent $y_i$ in a document $D$. The model is trained in an end-to-end manner, so every continuous span up to the maximum length is considered as a mention. Therefore, we work with $N = \frac{T(T+1)}{2}$ possible mentions, where $T$ is the number of words in a document $D$.

$P(y_i|D)$ is estimated from two scores: The first is the mention score of both mention $i$ and antecedent $y_i$, which express the likelihood of a span being a valid mention. The second is the binary antecedent score of $y_i$ being an antecedent of mention $i$. 

All three scores are summed together as follows:

\begin{equation}
    s(i, y_i) = 
    \begin{cases}
                0 & y_i = \epsilon \\
                s_m(i) + s_m(y_i) + s_a(i,y_i) & y_i \ne \epsilon\\  
    \end{cases},
\end{equation}\label{eq:scoring}
where $\epsilon$ is an empty (also referred as dummy) antecedent. If a mention candidate is aligned to the dummy antecedent and it is not an antecedent of any other mention (so the mention should be in the cluster with dummy antecedent only), it is not considered a valid mention. As a result, every valid mention has to be in a cluster with at least one other mention, so the model cannot predict singletons.

The scores are then normalized with the softmax function to estimate the probability:

\begin{equation}
    P(y_i|D) = \frac{\exp(s(i, y_i)}{\sum_{y' \in Y(i)}\exp(s(i, y')} 
\end{equation}

The scores are estimated by FFNN with one output neural without output activation as follows:

\begin{equation}
    s_m(i) = FFNN_m(g_i) 
\end{equation}

where $g_i$ is a vector representation of span $i$ obtained from text encoder (originally LSTM).

\begin{equation}
    s_a(i, j) = FFNN_a([g_i, g_j, g_i \odot g_j, \Phi]) 
\end{equation}

where $\Phi$ represents pairwise features lake distance between spans.

Vector representation $g_i$ of a span $i$ is optained from a concatenation of various features as follows:

\begin{equation}
    g_i = [x_{start(i)}, x_{end(i)}, x_i, \Phi]
\end{equation}

where $x$ is the whole input and $x_{start(i)}$, $x_{end(i)}$ are start and end token respectively. $x_i$ is the sum of all the span tokens weighted with the attention mechanism \cite{bahdanau2014neural}. $\Phi$ is an additional feature; span length.

In the training phase, we iterate all mentions predicted by a model and maximize the marginal probability of all its correct antecedents as shown in the following equation:

\begin{equation}
    J(D) = \log \prod_{i=1}^N \sum_{\hat{y} \in Y(i) \cap \texttt{GOLD}(i)}P(\hat{y})
\end{equation}\label{eq:loss}

where GOLD($i$) is the set of all gold antecedents of mention $i$ (in the training data annotation).

Theoretically, we should go over all possible mentions, but in practice, the mentions are prefiltered based on a unary mention score to reduce computational complexity.

\begin{figure*}
    \centering
    \begin{subfigure}{0.60\textwidth}
    \includegraphics[width=\textwidth]{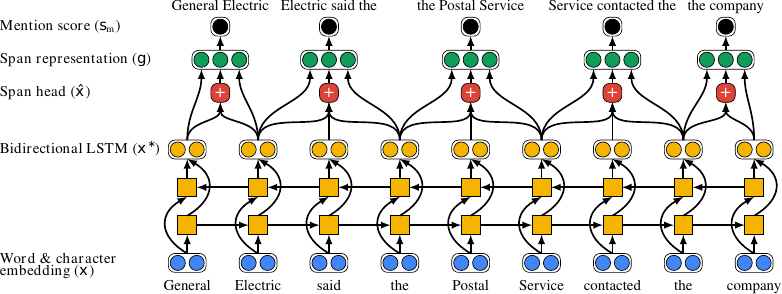}
    \caption{Mention Ranking}
    \label{fig:e2e_mention}
\end{subfigure}
\hfill
    \begin{subfigure}{0.38\textwidth}
    \includegraphics[width=\textwidth]{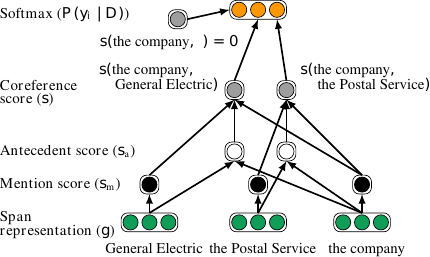}
    \caption{Antecedent Ranking}
    \label{fig:e2e_antecedent}
\end{subfigure}
\hfill
    
    \caption{Architecture of the model from \cite{lee-etal-2017-end}.}
    \label{fig:e2e_baseline}
\end{figure*}

The schema of the model is shown in Figure \ref{fig:e2e_baseline}.

This end-to-end model has proven to be highly successful, especially on datasets where singletons are not annotated (e.g., CoNLL 2012/OntoNotes), a scenario where traditional two-stage models tend to struggle. The model was later extended by \cite{lee2018higher} to the higher-order coreference resolution (CR) through iterative refining of span representations with gated attention mechanism. They further optimize pro model speed by pruning the mentions with simple scorer at the first step and rescoring only top $k$ spans with the precise scorer in the second step. Formally, they add the coarse coreference score: $S_c(i, j) = g_i \cdot w_c \cdot g_j$, which is used for antecedent pruning before computing the fine coreference score $s_a$. The method is called coarse-to-fine (c2f) coreference resolution.
Authors of \cite{joshi2019bert} use the model from \cite{lee2018higher} but with BERT as an encoder.

While many higher-order CR approaches have been proposed, authors of \cite{xu-choi-2020-revealing} demonstrated that their impact is marginal when a strong encoder is used, as seen in their experiments with SpanBERT \cite{joshi-etal-2020-spanbert}.

\subsection{Word-level Coreference Resolution}
Authors of \cite{dobrovolskii2021word} proposed a word-level coreference resolution approach based on the model by \cite{lee-etal-2017-end}. Their method reduces the computational complexity by reducing the mention space. Instead of iterating over all possible spans, they first map each gold mention to a single word—specifically, the headword, which they select based on the syntax tree (as the node closest to the root of the tree). Antecedent prediction is then performed on the word level. In the next step, they use a span extraction model to predict the original spans from headwords. This model is trained to classify each word in the sentence together with a headword to decide whether the word is the start token or end token of the span corresponding to the headword. During training, the span extraction model is trained simultaneously with the antecedent prediction as another classification head. The rest of the model is standard coarse-to-fine coreference resolution.

Later in \cite{d2023caw}, the authors proposed a modification in headword selection, suggesting that when a coordinate conjunction appears within an entity, it should be made the head to resolve span ambiguity. For example, in the sentence "Tom and Mary are playing," the entities "Tom" and "Tom, and Mary" would typically share the same head in the original approach. Their modification eliminates this ambiguity.


\subsection{CRAC Shared Task on Multilingual Coreference Resolution}

CRAC Shared Task on Multilingual Coreference Resolution (CRAC-coref) \cite{zabokrtsky-etal-2023-findings} is an annual shared task that began in 2022 and is built upon the CorefUD collection. In fact, this paper is an extension of CRAC22-coref \cite{crac2022findings} participant system \cite{prazak-konopik-2022-end}, which is based on \cite{pravzak2021multilingual}.   

\begin{figure}
    \centering
    \includegraphics[width=.7\linewidth]{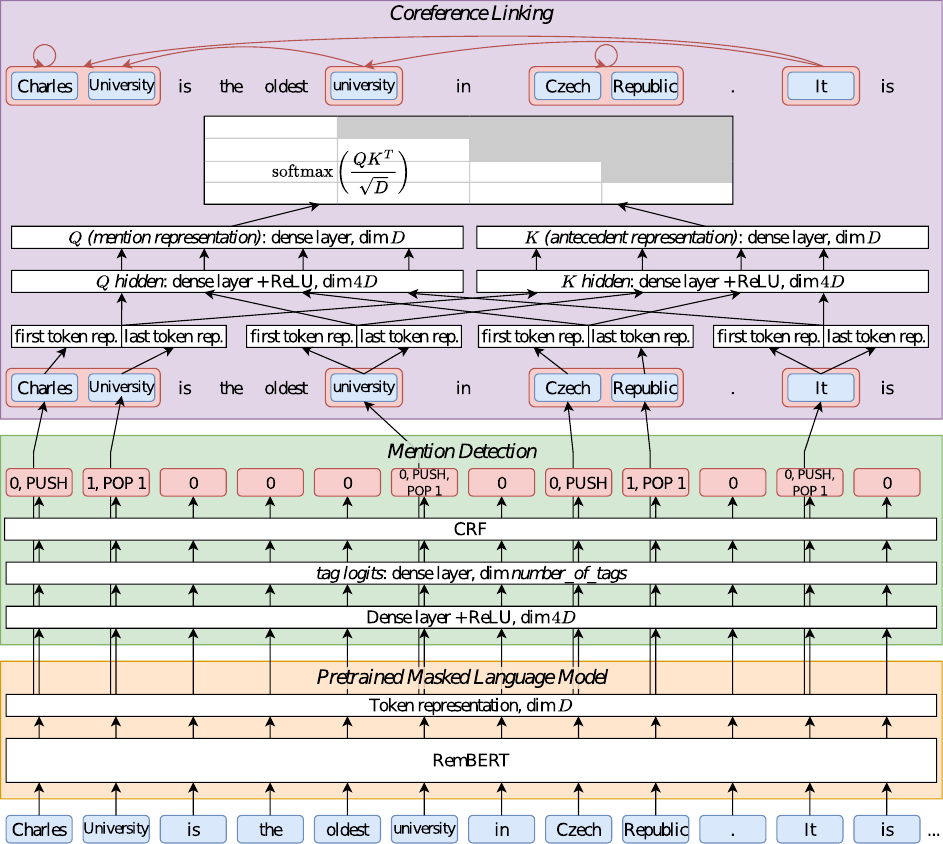}
    \caption{Schema of CorPipe model (from \cite{straka-strakova-2022-ufal})}
    \label{fig:corpipe}
\end{figure}

The best-performing system in CRAC 2022 was submitted by \cite{straka-strakova-2022-ufal}. It is a two-stage model with a joined BERT-like encoder. In the mention detection stage, they use extended BIO\footnote{BIO stands for Beginning, Inside, and Outside. It is a tagging format to label each token of a text as being the beginning (B), inside (I), or outside (O) of an entity.} tagging via stack manipulation instructions to solve overlapping mentions. Coreference resolution is then done by antecedent prediction head, a component responsible for identifying the most likely antecedent of each mention, which is very similar to the one used in end-to-end neural CR \cite{lee-etal-2017-end}. They use a concatenation of start and end tokens as mention representation and classify mention pairs to find the best antecedent of each mention. The schema of the CorPipe model is shown in Figure \ref{fig:corpipe}.

\section{Datasets \& Metrics}

For the evaluation of the proposed model, we use the CorefUD 1.1 dataset. The collection consists of 17 datasets in 12 different languages. All datasets in the collection are converted into a harmonized form.


The primary goal of CorefUD is to have as many datasets as possible in the same annotation scheme and enable cross-lingual training. Several important design choices have been made by the authors of CorefUD:

\begin{itemize}
    \item As the collection is the extension of Universal Dependencies \cite{nivre-etal-2020-universal}, all the datasets in the CorefUD collection are theebanks. Some of them were annotated in a treebank form (e.g., Czech-PDT), and the others were parsed during the harmonization procedure.
    \item Coreference entities are represented as clusters of mentions.
\end{itemize}

As highlighted in \cite{corefud2022lrec}, significant differences remain among individual datasets within CorefUD, not only in data distribution but also in annotation practices. In this section, we discuss the differences that are particularly relevant to our model and experiments.

\paragraph{Zero mentions and empty nodes:} Some datasets contain zero mentions nodes. These are mentions that are not explicitly mentioned in the text (cannot be mapped to a word or phrase), but they are implicitly referred to. In CorefUD, zero mantions are captured in empty nodes; virtual nodes are added to a syntactic tree during the annotation process, and they are not mapped to any word. Adding empty nodes as some special words can make it hard for a general-purpose pre-trained model to interpret the meaning of the text accurately. We cannot discard them from the datasets because most of them are part of a coreference cluster. The number of empty nodes in each dataset is shown in Table \ref{tab:datases_stats}, and the number of zero mentions can be found in Table \ref{tab:data_mantion_stats}.

\paragraph{Singletons:} Another critical difference between the datasets lies in the annotation of singletons. Singletons are entities with only a single mention throughout the document, so these mentions have no coreference relations. In some datasets, singletons are annotated, while in others, they are not. That means that when there is an entity with only a single mention, it is not considered an entity during annotation. Table \ref{tab:data_ent_stats} provides the percentage of singleton entities across the datasets (entities with a length of 1).

The CorefUD project is still evolving, and every year, there are some new datasets. The data is also moving towards a realistic scenario. In the very last version, 1.2, test data are practically just plain text (parsed with UDPipe, but no manual annotation is provided).

\begin{table}
    \centering
    \begin{adjustbox}{width=\linewidth}
    \begin{tabular}{@{}l rrrr r@{~~}r@{~}r@{}}\toprule
                         & \multicolumn{4}{c}{total size} & \multicolumn{3}{c}{division [\%]} \\ \cmidrule(lr){2-5}\cmidrule(l){6-8}
\pulrad{CorefUD dataset} &  docs &   sents &     words &   empty nodes & train &   dev & test \\ \midrule
Catalan-AnCora           &  1298 &  13,613 &   429,313 &   6,377 &  77.5 &  11.4 & 11.1 \\
Czech-PCEDT              &  2312 &  49,208 & 1,155,755 &  35,844 &  80.9 &  14.2 &  4.9 \\
Czech-PDT                &  3165 &  49,428 &   834,720 &  22,389 &  78.3 &  10.6 & 11.1 \\
English-GUM              &   195 &  10,761 &   187,416 &      99 &  78.9 &  10.5 & 10.6 \\
English-ParCorFull       &    19 &     543 &    10,798 &       0 &  81.2 &  10.7 &  8.1 \\
French-Democrat          &   126 &  13,057 &   284,883 &       0 &  80.1 &   9.9 & 10.0 \\
German-ParCorFull        &    19 &     543 &    10,602 &       0 &  81.6 &  10.4 &  8.1 \\
German-PotsdamCC         &   176 &   2,238 &    33,222 &       0 &  80.3 &  10.2 &  9.5 \\
Hungarian-KorKor         &    94 &   1,351 &    24,568 &   1,988 &  79.2 &  10.3 & 10.5 \\
Hungarian-SzegedKoref    &   400 &   8,820 &   123,968 &   4,857 &  81.1 &   9.6 &  9.3 \\
Lithuanian-LCC           &   100 &   1,714 &    37,014 &       0 &  81.3 &   9.1 &  9.6 \\
Norwegian-BokmaalNARC    &   346 &  15,742 &   245,515 &       0 &  82.8 &   8.8 &  8.4 \\
Norwegian-NynorskNARC    &   394 &  12,481 &   206,660 &       0 &  83.6 &   8.7 &  7.7 \\
Polish-PCC               &  1828 &  35,874 &   538,885 &     470 &  80.1 &  10.0 &  9.9 \\
Russian-RuCor            &   181 &   9,035 &   156,636 &       0 &  78.9 &  13.5 &  7.6 \\
Spanish-AnCora           &  1356 &  14,159 &   458,418 &   8,112 &  80.0 &  10.0 & 10.0 \\
Turkish-ITCC             &    24 &   4,733 &    55,341 &       0 &  81.5 &   8.8 &  9.7 \\
\bottomrule\end{tabular}
\end{adjustbox}
\caption{Basic dataset Statistics (taken from \cite{corefud2022lrec}).}
    \label{tab:datases_stats}
\end{table}

\begin{table}
    \centering
    \def\MC#1#2{\multicolumn{#1}{c}{#2}}
    \begin{adjustbox}{width=\linewidth}
    \begin{tabular}{@{}l rrrr rrrrr@{}}\toprule
                       & \MC{4}{Entities}                  & \MC{5}{distribution of lengths}      \\\cmidrule(lr){2-5}\cmidrule(l){6-10}
CorefUD dataset        &   total & per 1k & \MC{2}{length} &     1 &     2 &     3 &     4 &   5+ \\\cmidrule(lr){4-5}
                       &   count &  words &    max &  avg. &  [\%] &  [\%] &  [\%] &  [\%] & [\%] \\\midrule
Catalan-AnCora         &  18,030 &     42 &    101 &   3.5 &   2.6 &  54.0 &  18.1 &   8.6 &  16.8 \\
Czech-PCEDT            &  52,721 &     46 &    236 &   3.3 &   6.6 &  59.6 &  14.7 &   6.3 &  12.7 \\
Czech-PDT              &  78,747 &     94 &    175 &   2.4 &  40.8 &  35.1 &  10.4 &   4.9 &   8.9 \\
English-GUM            &  27,757 &    148 &    131 &   1.9 &  73.8 &  14.2 &   4.9 &   2.2 &   4.9 \\
English-ParCorFull     &     202 &     19 &     38 &   4.2 &   6.9 &  54.0 &  13.9 &   5.9 &  19.3 \\
French-Democrat        &  39,023 &    137 &    895 &   2.0 &  81.6 &  10.7 &   3.0 &   1.4 &   3.3 \\
German-ParCorFull      &     259 &     24 &     43 &   3.5 &   6.2 &  64.9 &  11.6 &   5.0 &  12.4 \\
German-PotsdamCC       &   3,752 &    113 &     15 &   1.4 &  76.5 &  13.9 &   5.0 &   1.8 &   2.7 \\
Hungarian-KorKor       &   1,134 &     46 &     41 &   3.6 &   0.9 &  55.1 &  17.1 &   9.0 &  17.9 \\
Hungarian-SzegedKoref  &   5,182 &     42 &     36 &   3.0 &   8.0 &  51.1 &  19.0 &   9.1 &  12.9 \\
Lithuanian-LCC         &   1,224 &     33 &     23 &   3.7 &  11.2 &  45.3 &  11.8 &   8.2 &  23.5 \\
Norwegian-BokmaalNARC  &  53,357 &    217 &    298 &   1.4 &  89.4 &   5.4 &   1.9 &   1.0 &   2.4 \\
Norwegian-NynorskNARC  &  44,847 &    217 &     84 &   1.4 &  88.7 &   5.5 &   2.1 &   1.1 &   2.7 \\
Polish-PCC             & 127,688 &    237 &    135 &   1.5 &  82.6 &   9.8 &   2.9 &   1.4 &   3.2 \\
Russian-RuCor          &   3,636 &     23 &    141 &   4.5 &   3.3 &  53.7 &  15.6 &   6.9 &  20.5 \\
Spanish-AnCora         &  20,115 &     44 &    110 &   3.5 &   3.3 &  53.7 &  17.0 &   8.8 &  17.1 \\
Turkish-ITCC           &     690 &     12 &     66 &   5.3 &   0.0 &  39.0 &  19.6 &  10.7 &  30.7 \\
\bottomrule\end{tabular}
\end{adjustbox}
\caption{Entity Statistics (taken from \cite{corefud2022lrec}). The entity length means the number of its mentions.}
    \label{tab:data_ent_stats}
\end{table}

\begin{table*}
    \centering
    \def\MC#1#2{\multicolumn{#1}{c}{#2}}
    \begin{adjustbox}{width=\linewidth}
\begin{tabular}{@{}l rrrr rrrrrr rrr@{}}\toprule

                       & \MC{4}{mentions}                  & \MC{6}{distribution of lengths}  & \MC{3}{mention type}            \\\cmidrule(lr){2-5}\cmidrule(l){6-11} \cmidrule(lr){12-14}
CorefUD dataset        &   total & per 1k & \MC{2}{length} &     0 &     1 &     2 &     3 &     4 &   5+ & w/empty & w/gap & non-tree \\\cmidrule(lr){4-5}
                       &   count &  words &    max &  avg. &  [\%] &  [\%] &  [\%] &  [\%] &  [\%] & [\%] & [\%] & [\%] & [\%] \\\midrule
Catalan-AnCora         &  62,417 &    145 &    141 &   4.8 &  10.2 &  28.2 &  21.7 &   7.9 &   5.3 &  26.8 &  12.4 &   0.0 &   3.7 \\
Czech-PCEDT            & 168,138 &    145 &     79 &   3.6 &  19.5 &  29.9 &  16.8 &   8.6 &   4.1 &  20.9 &  25.7 &   0.8 &   8.4 \\
Czech-PDT              & 154,983 &    186 &     99 &   3.1 &  10.7 &  39.6 &  20.3 &   9.1 &   4.2 &  16.1 &  13.7 &   1.3 &   2.2 \\
English-GUM            &  32,323 &    172 &     95 &   2.6 &   0.0 &  56.5 &  19.5 &   8.0 &   3.8 &  12.1 &   0.0 &   0.0 &   1.2 \\
English-ParCorFull     &     835 &     77 &     37 &   2.1 &   0.0 &  59.6 &  23.4 &   6.3 &   3.4 &   7.3 & 0.0 &   0.6 &  0.6 \\
French-Democrat        &  46,487 &    163 &     71 &   1.7 &   0.0 &  64.2 &  21.8 &   6.3 &   2.4 &   5.3 &   0.0 &   0.0 &   2.0 \\
German-ParCorFull      &     896 &     85 &     30 &   2.0 &   0.0 &  64.8 &  17.5 &   6.2 &   4.0 &   7.4 &   0.0 &   0.3 &   1.5 \\
German-PotsdamCC       &   2,519 &     76 &     34 &   2.6 &   0.0 &  34.8 &  32.4 &  15.6 &   6.4 &  10.9 &   0.0 &   6.3 &   3.8 \\
Hungarian-KorKor       &   4,103 &    167 &     42 &   2.2 &  30.8 &  20.6 &  23.2 &   9.8 &   4.8 &  10.8 &  35.3 &   0.6 &   5.5 \\
Hungarian-SzegedKoref  &  15,165 &    122 &     36 &   1.6 &  15.1 &  37.4 &  32.5 &  10.2 &   2.6 &   2.2 &  15.2 &   0.4 &   1.1 \\
Lithuanian-LCC         &   4,337 &    117 &     19 &   1.5 &   0.0 &  69.1 &  16.6 &  11.1 &   1.2 &   2.0 &   0.0 &   0.0 &   4.3 \\
Norwegian-BokmaalNARC  &  26,611 &    108 &     51 &   1.9 &   0.0 &  74.5 &   9.7 &   6.1 &   2.1 &   7.6 & 0.0 &   0.6 &   1.5 \\
Norwegian-NynorskNARC  &  21,847 &    106 &     57 &   2.1 &   0.0 &  70.2 &  10.1 &   7.7 &   2.8 &   9.2 &   0.0 &   0.4 &   1.4 \\
Polish-PCC             &  82,804 &    154 &    108 &   2.1 &   0.3 &  68.7 &  14.9 &   5.2 &   2.7 &   8.2 &   0.5 &   1.0 &   4.7 \\
Russian-RuCor          &  16,193 &    103 &     18 &   1.7 &   0.0 &  69.1 &  16.3 &   6.6 &   3.5 &   4.6 &   0.0 &   0.5 &   1.4 \\
Spanish-AnCora         &  70,663 &    154 &    101 &   4.8 &  11.4 &  31.6 &  18.8 &   7.2 &   4.5 &  26.3 &  14.0 &   0.0 &   0.3 \\
Turkish-ITCC           &   3,668 &     66 &     25 &   1.9 &   0.0 &  67.3 &  17.5 &   5.8 &   2.9 &   6.6 & 0.0 &   0.0 &   1.9 \\
\bottomrule\end{tabular}
\end{adjustbox}
\caption{Non-singleton mention statistics (taken from \cite{corefud2022lrec}). The mention length is the number of its non-empty nodes; thus, mentions with the length 0 contain only empty nodes. \textit{w/empty} are mentions where at least one empty node is present e., g., zero mentions. \textit{w/gap} are mentions with at least one gap, so these are mentions which cannot be represented with a single span. \textit{non-tree} are mentions which do not form a single subtree of the syntax tree.}
    
    \label{tab:data_mantion_stats}
\end{table*}

\subsection{Evaluation Metrics}

For evaluation, we employ the official metric for CRAC 2023 Multilingual Coreference Resolution shared task implemented via the CoreUD scorer\footnote{\url{https://github.com/ufal/corefud-scorer}}. The metric is a modification of the CoNLL score for coreference resolution, averaging three F1 metrics: \textit{CEAF-e}, $B^3$, and \textit{MUC}. Each of these three metrics has a different universe. \textit{MUC} operates on the level of coreference links, $B^3$ works with mentions and \textit{CEAF-e} is entity-based metric. All these three metrics are described later in this section. The main modification of the CorefUD scorer lies in using the head matching. In head matching, the gold and system mentions are considered identical if and only if they have the same syntactic head. Another important aspect is that the primary metric ignores singletons.

\subsubsection{\textit{MUC} Score}
The \textit{MUC} score \cite{vilain1995model} is a link-based metric. It measures precision and recall of the identification of coreference links in a coreference chain. Coreference chains are formed by linking each mention in a cluster to its antecedent, creating a sequence of references. For instance, precision is computed by summing all correctly predicted links divided by all the links in the system response. In recall, the numerator is the number of all the gold links. Then, we compute the standard F1 measure as the final metric. 
The number of correctly identified links is technically computed from a partition function $\mathcal{P}(G_{i}$, which determines how many entities cover all the mentions of gold entity $G_i$. Recall is then computed as: 
\begin{equation}
     \text{Recall}_{\text{MUC}} = \frac{\sum_{i} |G_{i}| - |\mathcal{P}(G_{i}, R)|}{\sum_{i} |G_{i}| - 1}
 \end{equation}

\subsubsection{$\text{B}^3$}

$B^3$ \cite{bagga1998algorithms} was proposed to address problems of \textit{MUC} score. One of them is that \textit{MUC} score as a link-based metric does not address singletons at all. $B^3$ operates on mention level. It computes precision and recall for every mention and then averages these values. the recall and precision of a mention $m$ is defined as follows:

\begin{equation}
    R_m = \frac{|G_{i} \cap S_{j}|}{|G_{i}|}
\end{equation}

\begin{equation}
    P_m = \frac{|G_{i} \cap S_{j}|}{|S_{i}|}
\end{equation}

\subsubsection{\textit{CEAF}}

The problem with $B^3$ is that each entity is addressed many times (with each mention). CEAF \cite{luo2005coreference} tries to solve this problem. It aligns gold and system entities one-to-one, maximizing the similarity of aligned entities. This leads to the maximum bipartite matching problem. 
Precision and recall are then defined as follows: 

\begin{equation}
    P = \frac{\Phi(g^*)}{\sum_i\phi(S_i, S_i)}
\end{equation}

where $\Phi(g^*)$ is sum of similarities of all the matching entities in one-to-one mapping found by Kuhn-Munkres Algorithm, and $\phi(x, x)$ is self-similarity of an entity.

\cite{luo2005coreference} present two entity-pair similarity measures that can be used in \textit{CEAF}: one is the absolute number of common mentions between two entities, and the other is a “local” mention F-measure between two entities. The two measures lead to
the mention-based \textit{CEAF-m} and entity-based \textit{CEAF-e}, respectively. \cite{poesio2023computational}

\section{Baseline Model \& Extensions}
\label{sec:extensions}
We adopt an end-to-end model from \cite{xu-choi-2020-revealing}, and we use it as the baseline model (without higher-order inference). We use UDAPI\footnote{\url{https://github.com/udapi/udapi-python}} to load the data in CorefUD format. We use XLM-Roberta-large as an encoder.

 Building on this baseline, we propose several extensions to the standard end-to-end model presented earlier in this paper. Our objective is to create a universal model suitable for all datasets in the CorefUD collection. We also reevaluate some extensions proposed by \cite{prazak-konopik-2022-end} for comparison (namely syntax representation and \textit{Span2head} model).

\subsection{Cross-lingual Training}
\label{sec:joined_model}
Some of the datasets are relatively small (see Table \ref{tab:datases_stats}). To enhance performance, we propose pretraining the model on a concatenation of all training datasets within the CorefUD collection.

\begin{table}
    \centering
    \begin{tabular}{lcc}
    \toprule
         Model & Pretrained params & New params  \\
         \midrule
         mBERT & 180M & 15M \\
         XLM-R & 550M & 20M \\
         \bottomrule
    \end{tabular}
    \caption{Number of trainable parameters of the models}
    \label{tab:num_params}
\end{table}


Table \ref{tab:num_params} shows the total number of parameters and the number of parameters trained from scratch. We can see that XLM-R large has approximately 20 million parameters trained from scratch\footnote{The number is the average of different configurations since different extensions add different numbers of parameters.}. Training so many parameters from scratch can lead to overfitting, especially for smaller datasets. To mitigate this, we first train a joined model on the concatenation of all the datasets and then we finetune this model on each dataset separately.

\subsection{Syntactic Information}
We believe that integrating syntactic information can greatly improve the performance. Dependency information is crucial for identifying mention heads, as some datasets in the CorefUD collection label mentions as nodes within the dependency tree rather than spans in the linear text sequence.

We use the syntactic information model from \cite{prazak-konopik-2022-end}. In this model, we modify a model to encode syntactic information in the following way: For each token, we add its path to \textit{ROOT} to its embedding representation. To encode a path in a dependency tree, we use a concatenation of BERT embeddings of all the nodes on the path together with the learned embedding of corresponding dependency relations. To have a fixed-length word representation, we set a maximum tree-path length. By adding tree representation, we increase the total size of a token representation in the model by $max\_tree\_depth \times (bert\_emb\_size + deprel\_emb\_size)$. In our experiments, we use the maximum tree-path length of 5.


\subsection{Span2head Model} \label{sec:extension_span2head}

Since the official evaluation uses a head-match metric, we trained the model to predict the heads of mentions rather than full spans to better align with the evaluation criterion. With access to all the necessary information, including dependency trees, the model should be able to infer the original rules for selecting heads.

Our goal is to represent mention with a headword only. However, multiple mentions can share the same head. To address this issue, \textit{Span2head} model represents mentions by their full spans and predicts the head of each mention at the top of the CR model, outputting only the headword(s). This ensures that, during cluster formation, mentions are represented by their spans, preventing incorrect merging of clusters for distinct mentions that share a headword.

\begin{figure}
    \centering
    \includegraphics[width=0.7\linewidth]{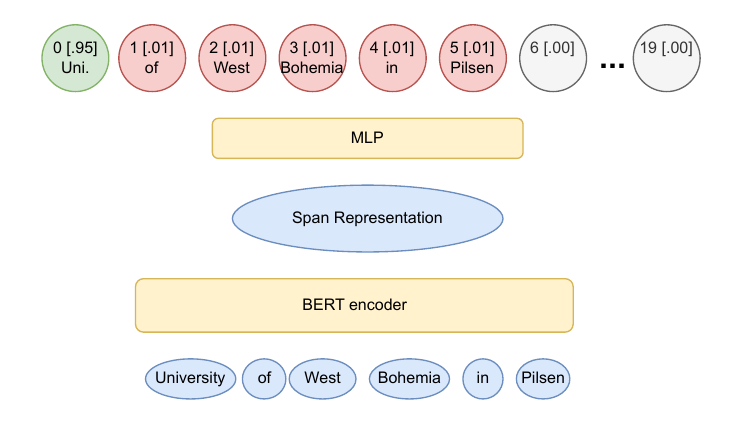}
    \caption{Multi-class Span2Head Model.}
    \label{fig:span2head_multi}
\end{figure}

Authors of \cite{prazak-konopik-2022-end} proposed two versions of the head prediction model, both implemented as separate classification layers on top of our coreference resolution model. The first \textit{multi-class} model predicts the relative position of the headword(s) within a span, using the span's hidden representation from the coreference model. Head position probabilities are calculated with sigmoid activation, allowing the model to predict multiple headwords, even though only one is present in the gold data. This serves to optimize the evaluation metric by outputting all likely headword candidates. The schema of the model is shown in Figure \ref{fig:span2head_multi}

\begin{figure}
    \centering
    \includegraphics[width=0.7\linewidth]{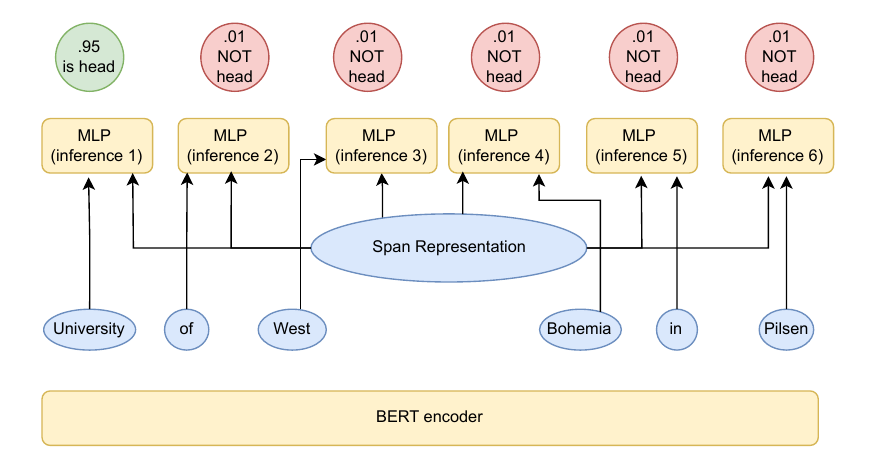}
    \caption{Binary Span2Head Model.}
    \label{fig:span2head_binary}
\end{figure}

The second \textit{binary} model employs binary classification for each span and head candidate pair, which also enables the prediction of multiple headwords for a single span. The schema of the model is shown in Figure \ref{fig:span2head_binary}.

\subsection{Head Representations}

Inspired by the word-level coreference resolution model, we decided to try to model mentions only with their syntactic head. As mentioned earlier, this model merges some mentions into one in cases where multiple mentions share the same head. However, according to \cite{dobrovolskii2021word}, reduction to headword is highly beneficial. As also mentioned above, for the CRAC official evaluation metric, reconstructing the original span is unnecessary, as predicting the correct head suffices. Therefore, we explored a simplified word-level coreference resolution approach that bypasses the span prediction step. Given that syntactic information is available in all CorefUD datasets, this is not an unrealistic scenario because, in most cases, the spans can be reconstructed by simply taking the whole subtree of the predicted node (head). Heads are heuristically selected only in cases where a mention does not form a single subtree. The proportion of such mentions is shown in Table \ref{tab:data_mantion_stats}(column \textit{non-tree}). Using a word-level model reduces the mention space from quadratic to linear, making the model more efficient and minimizing potential false-positive mentions. Moreover, we believe that for very long mentions, the standard representation (sum of the start token, end token, and attended sum of all tokens) becomes insufficient.

\subsection{Singletons}

Some datasets in the CorefUD collection have singletons annotated, and others do not. Specifically, in CorefUD 1.1, 8 out of 17 datasets have more than 10\% singletons, and 6 of these have more than 70\% singletons, which is probably a sign of consistent entity annotation independent of the coreference annotation. The baseline model completely ignores singletons during training\footnote{The loss is the sum over all correct antecedents, and since singletons have no gold antecedents, they do not affect the loss.}. As a result, for these 6 singleton-including datasets, we discard more than 70\% of training data for the mention identification task. To leverage this data, we incorporate singleton modeling into our model and propose several approaches to address this.

\subsubsection{Another Dummy Antecedent}

In the first proposed approach, we introduce another virtual antecedent (with trained embedding) representing that the mention has no real antecedent, but it is a valid mention, so it is a singleton. For these mentions, we avoid using the binary score because it would unnecessarily model the similarity between all the singletons \footnote{because all the singleton representations would be trained to be similar to the dummy antecedent}. Additionally, we include an option to use a separate feed-forward network for computing singleton scores (different from the one for other mentions scores but trained simultaneously on the top of the same encoder).

\subsubsection{Mention Modeling}
\label{sec:extension_mention}
The second approach modifies the loss function to model mentions independently of coreference relations. In this approach, we simply add a binary cross-entropy of each span being a mention to the loss function. In other words, we add another classification head for the mention classification:

\begin{equation}
    J(D) = \log \prod_{i=1}^N \sum_{\hat{y} \in Y(i) \cap \texttt{GOLD}(i)}P(\hat{y}) + \underbrace{y_m^{(i)} \cdot \sigma(s_m(i)) + (1- y_m^{(i)}) \cdot \sigma(-s_m(i))}_\text{singletons binary cross-entropy}
\end{equation}
where $y_m^{(i)}$ is $1$ if span $i$ corresponds to gold mention, $0$ otherwise.

In the prediction step, the mention score is evaluated only for potential singletons. If a mention has no real antecedent, we look at the mention score. If it is likely to be a mention, we make it a singleton; otherwise, it is not a mention at all.

\subsection{Overlapping Segments} \label{sec:overlapping}

One limitation of the XLM-R model is the short sequence length. In the employed model, the input document is split into segments, which are processed individually with a BERT-like encoder and merged in the antecedent-prediction head. To propagate errors correctly across segments, we need to process all segments in a single gradient update. However, due to GPU memory constraints, we set a maximum number of segments, and if a document exceeds this limit, we split it into multiple individual documents without any mutual coreferences.

\begin{figure}
    \centering
    \includegraphics[width=0.4\linewidth]{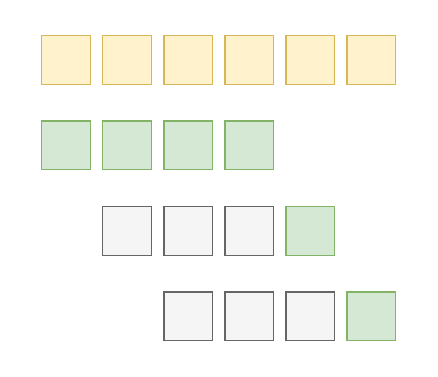}
    \caption{Example of overlapping segments splitting.}
    \label{fig:overlapping_segments}
\end{figure}

We propose using segment overlapping with a cluster merging algorithm to address this issue. The cluster merging algorithm is straightforward: We iterate over all mentions in the new part, and if it is present in a cluster in the previous part, we simply make a union between its clusters from both parts. We experiment with minimal (just one overlapping segment) and maximal (each continuing example has just one new segment the rest is overlapping) segment overlap. For instance, if a document has 6 segments and the maximum number of segments for a single example is 4, then we split the document into 3 examples overlapping with 3 segments. The example is shown in Figure \ref{fig:overlapping_segments}. We further extend this method by only using clusters in the continuing examples where at least one mention is present in new segments. If there is no such mention then all the mentions were seen already in the previous part with a longer left context. 

\section{Experiments}

\subsection{Training}

We trained our models using NVIDIA A40 GPUs. the batch size was equal to one document. We limit the maximum sequence length to 6 non-overlapping segments of 512 tokens. During training, if a document exceeded this length, a random offset was used to select a continuous block of 6 segments, with the remaining segments discarded. During prediction, longer documents are split into independent sub-documents. In the basic model, there is no overlap between sub-documents. In the case of the head-only model, we increased the number of segments to eight since this model is more memory-efficient, allowing for a larger sequence length. Each model was trained on its respective dataset for around 80,000 updates in our monolingual experiments. For the models pre-trained on multiple datasets, we used 80,000 steps for joint pre-training across all datasets, followed by roughly 30,000 steps for fine-tuning on each individual dataset.

\subsection{Baseline and Extensions}

We begin by training a monolingual model for each language. For these experiments, we use XLM-Roberta-large and specific monolingual models for each language. Monolingual models for individual languages are listed in Table \ref{tab:base_results}. Then, we train a joined cross-lingual model for all the datasets, which we fine-tune for each dataset afterward.

To evaluate the effect of each extension, we first run all its variants described in \ref{sec:extensions} on the basic XLM-R-large model and measure the performance for each language (Sections \ref{sec:disscussion_span2head} and \ref{sec:disscussion_singletons}). Then, we compare each extension in its best configuration to the baseline joined model in Section \ref{sec:discussion_best} (since joined models achieve the best results from our three baseline models). In the last step, we evaluate the effects of Overlapping segments only on the best model configuration for each language (because it is orthogonal to all other extensions).

\subsubsection{Monolingual Models}

We evaluate a specific monolingual model for each language. We use a large variant where it is available. For Czech we use \textit{Czert} \cite{sido2021czert}, for English \textit{Roberta-large} \cite{zhuang2021robustly}, gbert-large \cite{chan2020german} for German. 
The complete list can be found in Table \ref{tab:base_results}.

\subsection{Zero-shot Cross-lingual Experiments}

We also conduct zero-shot cross-lingual experiments in two variants: dataset zero-shot and language zero-shot. In dataset zero shot, we train the model on all the datasets except one, and then we evaluate the model on the excluded dataset. However, this approach is not a true cross-lingual zero-shot test, as multiple datasets exist for some languages, meaning the model was still exposed to the language of the evaluation data during training.

To address this, we performed language zero-shot experiments, where we removed all datasets for a particular language from the training data. Then, we evaluated the model on all datasets in that language. Additionally, we conducted a final experiment focusing on the \textit{ParCor} corpora, which consists of parallel English and German datasets. This allowed us to perform a true zero-shot evaluation on both \textit{ParCor} corpora.

\section{Results \& Discussion}

Table \ref{tab:base_results} presents the results of baseline models on development datasets parts. Several key observations can be drawn from these results. Most notably, monolingual models tend to outperform the cross-lingual XLM-Roberta model, especially for the Lithuanian and Czech PCEDT datasets, where the monolingual models—even though they are smaller—exhibit significantly better performance. The reason is probably in the difference of these two datasets from the other ones. We can see that for these two datasets, joined training helps only slightly compared to other datasets. For the rest of the datasets, joined training surpasses both other models (monolingual and multilingual trained from scratch).

\begin{table*}
    \centering
    \begin{adjustbox}{width=\linewidth,center}

\begin{tabular}{@{}lllllr@{}}
\toprule
Dataset/Model                        & monolingual model name             & reference                                                          & \multicolumn{1}{c}{Monoling} & XLM-R                 & \multicolumn{1}{l}{joined} \\ \midrule
\multicolumn{1}{l|}{ca\_ancora}      & PlanTL-GOB-ES/roberta-base-ca      & \multicolumn{1}{l|}{\cite{armengol-estape-etal-2021-multilingual}} & \textbf{74.23 $\pm$ .55}     & 70.51 $\pm$ .65       & \textbf{74.42 $\pm$ .25}   \\
\multicolumn{1}{l|}{cs\_pcedt}       & Czert-B-base-cased                 & \multicolumn{1}{l|}{\cite{sido2021czert}}                          & \textbf{73.85 $\pm$ .24}     & 72.28 $\pm$ .28       & {\ul 73.23 $\pm$ .08}      \\
\multicolumn{1}{l|}{cs\_pdt}         & Czert-B-base-cased                 & \multicolumn{1}{l|}{\cite{sido2021czert}}                          & 70.29 $\pm$ .35              & {\ul 71.87 $\pm$ .7}  & \textbf{73.47 $\pm$ .23}   \\
\multicolumn{1}{l|}{de\_parcorfull}  & deepset/gbert-base                 & \multicolumn{1}{l|}{\cite{chan2020german}}                         & 59.83 $\pm$ 2                & {\ul 71.79 $\pm$ 3.6} & \textbf{76.57 $\pm$ 1.5}   \\
\multicolumn{1}{l|}{de\_potsdamcc}   & deepset/gbert-large                & \multicolumn{1}{l|}{\cite{chan2020german}}                         & 64.78 $\pm$ 1.7              & {\ul 68.98 $\pm$ 2.4} & \textbf{76.3 $\pm$ 1.1}    \\
\multicolumn{1}{l|}{en\_gum}         & roberta-large                      & \multicolumn{1}{l|}{\cite{zhuang2021robustly}}                     & {\ul 67.415 $\pm$ 2.5}       & 65.73 $\pm$ .36       & \textbf{69.04 $\pm$ .49}   \\
\multicolumn{1}{l|}{en\_parcorfull}  & roberta-large                      & \multicolumn{1}{l|}{\cite{zhuang2021robustly}}                     & 66.39 $\pm$ 7.0              & 67.66 $\pm$ 2.9       & 69.16 $\pm$ 2.0            \\
\multicolumn{1}{l|}{es\_ancora}      & PlanTL-GOB-ES/roberta-large-bne    & \multicolumn{1}{l|}{\cite{gutierrezfandino2022}}                   & 70.73 $\pm$ .23              & {\ul 73.27 $\pm$ .7}  & \textbf{76.09 $\pm$ .13}   \\
\multicolumn{1}{l|}{fr\_democrat}    & camembert/camembert-large          & \multicolumn{1}{l|}{\cite{martin2020camembert}}                    & 57.08 $\pm$ .9               & 57.35 $\pm$ 1.6       & \textbf{65.24 $\pm$ .49}   \\
\multicolumn{1}{l|}{hu\_korkor}      & SZTAKI-HLT/hubert-base-cc          & \multicolumn{1}{l|}{\cite{nemeskey2021a}}                          & 60.68 $\pm$ 1.6              & 59.44 $\pm$ 1.0       & \textbf{68.99 $\pm$ .95}   \\
\multicolumn{1}{l|}{hu\_szegedkoref} & SZTAKI-HLT/hubert-base-cc          & \multicolumn{1}{l|}{\cite{nemeskey2021a}}                          & 66.75 $\pm$ 1.7              & 67.67 $\pm$ .44       & \textbf{69.52 $\pm$ .59}   \\
\multicolumn{1}{l|}{lt\_lcc}         & EMBEDDIA/litlat-bert               & \multicolumn{1}{l|}{\cite{ulvcar2021training}}                     & \textbf{76.84 $\pm$ 1.1}     & 72.52 $\pm$ .48       & {\ul 73.59 $\pm$ 1.1}      \\
\multicolumn{1}{l|}{no\_bokmaalnarc} & ltg/norbert3-large                 & \multicolumn{1}{l|}{\cite{samuel-etal-2023-norbench}}              & {\ul 73.47 $\pm$ 1.2}        & 72.11 $\pm$ .78       & \textbf{74.79 $\pm$ .44}   \\
\multicolumn{1}{l|}{no\_nynorsknarc} & ltg/norbert3-large                 & \multicolumn{1}{l|}{\cite{samuel-etal-2023-norbench}}              & {\ul 73.51 $\pm$ .67}        & 72.32 $\pm$ .55       & \textbf{75.98 $\pm$ .17}   \\
\multicolumn{1}{l|}{pl\_pcc}         & allegro/herbert-large-cased        & \multicolumn{1}{l|}{\cite{mroczkowskietal2021herbert}}             & \textbf{74.3 $\pm$ .29}      & 72.48 $\pm$ .61       & \textbf{74.1 $\pm$ .23}    \\
\multicolumn{1}{l|}{ru\_rucor}       & DeepPavlov/rubert-base-cased       & \multicolumn{1}{l|}{\cite{kuratov2019adaptation}}                  & 64.69 $\pm$ .65              & {\ul 68.52 $\pm$ .35} & \textbf{70.38 $\pm$ .64}   \\
\multicolumn{1}{l|}{tr\_itcc}        & dbmdz/electra-base-tur.-cased-dis. & \multicolumn{1}{l|}{}                                              & 18.39 $\pm$ 1.3              & {\ul 21.49 $\pm$ 1.6} & \textbf{45.39 $\pm$ 1.3}   \\ \midrule
avg                                  &                                    &                                                                    & 65.48                        & 66.23                 & \multicolumn{1}{l}{70.96}  \\ \bottomrule
\end{tabular}
\end{adjustbox}
\caption{Results of baseline models on dev data. CRAC 2023 official evaluation metric. \textit{Monoling} column shows the result of the monolingual model specific to each language. The \textit{XLM-R} column presents results of XLM Roberta large trained separately for each dataset. The \textit{Joined} column corresponds to the joined model described in Section \ref{sec:joined_model}.}
    \label{tab:base_results}
\end{table*}

\subsection{Effects of Extensions}

\subsubsection{Joined Training}

We further investigate the effect of joined training on the Czech PDT dataset by analyzing concrete documents. We found that the joined model is much better in identifying mentions. For example, if we look at the following part of a document with the gold annotations:

\begin{description}
    \item[GOLD] \textit{\textcolor{red}{$\|$}\textcolor{blue}{Visegrádská$_{e2}$} \textcolor{red}{trojka$\|_{e1}$} zřejmě nebude mít \textcolor{green}{volný obchod$_{e3}$}. \textcolor{cyan}{$\|$}\textcolor{magenta}{$\|$}\textcolor{orange}{Maďarsko$_{e6}$} \textcolor{magenta}{a} \textcolor{violet}{Polsko$_{e7}$}\textcolor{magenta}{$\|_{e5}$} \textcolor{cyan}{se rozhodly ustoupit od chystané zóny} \textcolor{green}{$\|$volného obchodu mezi zeměmi} \textcolor{red}{$\|$}\textcolor{blue}{visegrádské$_{e2}$} \textcolor{red}{trojky$\|_{e1}$}\textcolor{green}{$\|_{e3}$}\textcolor{cyan}{$\|_{e4}$}}
    \item[GOLD in English] \textit{\textcolor{red}{$\|$}\textcolor{blue}{The Visegrad$_{e2}$} \textcolor{red}{Three$\|_{e1}$} will probably not have \textcolor{green}{free trade$_{e3}$}. \textcolor{cyan}{$\|$}\textcolor{magenta}{$\|$}\textcolor{orange}{Hungary$_{e6}$} \textcolor{magenta}{and} \textcolor{violet}{Poland$_{e7}$}\textcolor{magenta}{$\|_{e5}$} \textcolor{cyan}{have decided to withdraw from the planned} \textcolor{green}{$\|$free trade zone between} \textcolor{red}{$\|$}\textcolor{blue}{Visegrad$_{e2}$} \textcolor{red}{Three$\|_{e1}$}\textcolor{green}{$\|_{e3}$}\textcolor{cyan}{$\|_{e4}$}}
    \item[Baseline] \textit{\textcolor{red}{$\|$}\textcolor{blue}{Visegrádská$_{e2}$} \textcolor{red}{trojka$\|_{e1}$} zřejmě nebude mít volný obchod. \textcolor{green}{Maďarsko$_{e3}$} a Polsko se rozhodly ustoupit od chystané zóny volného obchodu mezi zeměmi \textcolor{red}{$\|$}\textcolor{blue}{Visegrádské$_{e2}$} \textcolor{red}{trojky$\|_{e1}$}.}
    \item[Joined] \textit{"\textcolor{red}{$\|$}\textcolor{blue}{Visegrádská$_{e2}$} \textcolor{red}{trojka$\|_{e1}$} zřejmě nebude mít \textcolor{green}{volný obchod$_{e3}$}. \textcolor{magenta}{$\|$}\textcolor{orange}{Maďarsko$_{e6}$} \textcolor{magenta}{a} \textcolor{violet}{Polsko$_{e7}$}\textcolor{magenta}{$\|_{e5}$} se rozhodly ustoupit od chystané zóny \textcolor{green}{$\|$volného obchodu mezi zeměmi} \textcolor{red}{$\|$}\textcolor{blue}{visegrádské$_{e2}$} \textcolor{red}{trojky$\|_{e1}$}\textcolor{green}{$\|_{e3}$}"},
    
    where only the very long entity \textit{e4} is missing. (This is caused by maximum mention length, as discussed later).
\end{description}

 We have also noticed that the most significant difference between the baseline and joined model is in the case of foreign news, where the same topic can appear in multiple datasets (as this news from the European parliament)

\subsubsection{Span2Head}
\label{sec:disscussion_span2head}

Table \ref{tab:s2h_results} shows the performance of different variants of \textit{Span2head} model. We added \textit{Heads} column with heads-only mention representation to show if it is better to use the Span2head model or the heads-only model (since they are mutually exclusive). From the results, we can make several conclusions: 
\begin{enumerate}
    \item Span2head model slightly helps.
    \item The Binary model achieves better results than the multi-class model.
    \item For all the datasets where the Span2head model helps, the heads-only model is much better, so it does not make sense to use the Span2head model in the next experiments.
\end{enumerate}

Another interesting point is that for a monolingual model trained from scratch, the heads-only model improves the results on 13 out of 17 datasets, and the results are worse only for a single dataset (English-ParCor), which is very small, so the difference might be caused by random chance.

\begin{table}
    \centering
\begin{tabular}{@{}llrr|l@{}}
\toprule
Dataset/S2H model & None           & \multicolumn{1}{l}{Multi-class} & \multicolumn{1}{l|}{Binary} & Heads          \\ \midrule
ca\_ancora        & {\ul 70.7}     & 65.08                     & 65.16                       & \textbf{78.93} \\
cs\_pcedt         & 72.06          & 62.73                     & 69.77                       & \textbf{74.83} \\
cs\_pdt           & 71.21          & 66.19                     & {\ul 71.64}                 & \textbf{78.08} \\
de\_parcorfull    & \textbf{72.99} & 64.44                     & 65.73                       & \textbf{73.18} \\
de\_potsdamcc     & 68.26          & 66.03                     & {\ul 70.77}                 & \textbf{73.28} \\
en\_gum           & 64.06          & 61.13                     & {\ul 64.64}                 & \textbf{72.95} \\
en\_parcorfull    & \textbf{69.65} & 58.4                      & 64.42                       & 67.42          \\
es\_ancora        & 73.18          & 66.75                     & 65.2                        & \textbf{79.7}  \\
fr\_democrat      & 55.97          & 50.18                     & 51.3                        & \textbf{60.78} \\
hu\_korkor        & \textbf{60.08} & 46.52                     & 58.32                       & \textbf{60.34} \\
hu\_szegedkoref   & 68.02          & 63.76                     & 67.44                       & \textbf{68.87} \\
lt\_lcc           & 72.48          & 69.5                      & 72.01                       & \textbf{75.13} \\
no\_bokmaalnarc   & \textbf{72.83} & 67.52                     & 72.31                       & \textbf{72.93} \\
no\_nynorsknarc   & 72.2           & 67.3                      & 71.34                       & \textbf{75.05} \\
pl\_pcc           & 72.32          & 69.97                     & {\ul 72.39}                 & \textbf{74.19} \\
ru\_rucor         & 68.68          & 66.03                     & {\ul 69.42}                 & \textbf{72.05} \\
tr\_itcc          & 20.51          & 17.32                     & 17.08                       & \textbf{27.82} \\ \hline
avg               & 66.19                    & 60.52                     & 64.06                       & 69.74    \\ \bottomrule
\end{tabular}
\caption{Results of XLM-R baseline with different types of Span2Head model. \textit{None} column shows the results of the baseline XLM-R-large model without any extensions.}
    \label{tab:s2h_results}
\end{table}

\subsubsection{Singletons}
\label{sec:disscussion_singletons}

Table \ref{tab:singletons_results} shows the results of all proposed variants of the singletons extension. \textit{None} column represents the baseline \textit{xlm-r-large} trained from scratch for each dataset. \textit{Dummy} column shows the results of plain dummy antecedent for singletons. \textit{mask} extends the previous method binary score masking for singletons. This means that the method does not use the binary score for similarities to the singleton dummy antecedent. \textit{separate} uses separate FFNN for singletons score prediction (different from the one used for standard mentions). The last method \textit{mentions} represents the method described in section \ref{sec:extension_mention}. Additionally, the Table shows the number of parameters added by each variant of the model. The last column shows the percentage of singleton entities in each dataset (datasets with few singletons do not need any singleton model).
\begin{table*}
    \centering
\scalebox{0.85}{
\begin{tabular}{lrrrrr|r}
\hline
Model           & None                      & Dummy                     & Mask                      & Separate                  & Mentions                   & \% of singletons \\
+params         & \multicolumn{1}{l}{0}     & \multicolumn{1}{l}{3092}  & \multicolumn{1}{l}{3092}  & 40M                       & 40M                        & -                \\ \hline
ca\_ancora      & 70.7                      & 70.81                     & 68.75                     & 70.82                     & \textbf{71.61}             & 2.6              \\
cs\_pcedt       & \textbf{72.06}            & \textbf{72.03}            & 71.65                     & 71.81                     & \textbf{72.44}             & 6.6              \\
cs\_pdt         & 71.21                     & 70.75                     & 70.04                     & 72.23                     & \textbf{72.84}             & 40.8             \\
de\_parcorfull  & \textbf{72.99}            & 70.23                     & 70.09                     & 65.47                     & 71.29                      & 6.2              \\
de\_potsdamcc   & 68.26                     & 67.99            & 68.24                     & 68.44                     & 67.49                      & 76.5             \\
en\_gum         & 64.06                     & 70.4                      & 69.81                     & 70.89                     & \textbf{71.61}             & 73.8             \\
en\_parcorfull  & \textbf{69.65}            & 67.49                     & 64.78                     & 65.89                     & 64.43                      & 6.9              \\
es\_ancora      & 73.18                     & 72.88                     & 70.98                     & \textbf{73.74}            & \textbf{73.85}             & 3.3              \\
fr\_democrat    & 55.97                     & 55.6                      & 53.96                     & 53.86                     & \textbf{59.75}             & 81.6             \\
hu\_korkor      & \textbf{60.08}            & \textbf{60.56}            & 56.6                      & 55.59                     & 57.91                      & 0.9              \\
hu\_szegedkoref & \textbf{68.02}            & 67.1                      & 66.57                     & 67.1                      & 67.03                      & 8.0              \\
lt\_lcc         & 72.48                     & 72.15                     & \textbf{75.59}            & 73.3                      & 74.48                      & 11.2             \\
no\_bokmaalnarc & 72.83                     & 73.28                     & 73.33                     & 73.84                     & \textbf{74.15}             & 89.4             \\
no\_nynorsknarc & 72.2                      & 72.99                     & 73.57                     & \textbf{74.77}            & 74.26                      & 88.7             \\
pl\_pcc         & 72.32                     & 73.78                     & 72.81                     & \textbf{74.4}             & \textbf{74.49}             & 82.6             \\
ru\_rucor       & \textbf{68.68}            & 67.56                     & 67.91                     & 68.55                     & \textbf{69.12}             & 3.3              \\
tr\_itcc        & \textbf{20.51}            & \textbf{21.11}            & 16.07                     & 20.24                     & 13.21                      & 0.0              \\ \hline
avg             & 66.19 & 66.38 & 65.34 & 65.94 & 66.47 & -                 \\ \hline
\end{tabular}
}
\caption{Results of XLM-R baseline with different types of singletons model}
    \label{tab:singletons_results}
\end{table*}

The table shows that, on average, the best results were achieved with the mention model. \textit{Mask} model has very poor results. Still, it makes sense because the number of specific parameters for singletons modeling is very low (3 092 params are used as embedding for dummy antecedent). If we discard the binary score, we reduce the learning power of the model even more. \textit{Dummy} model does not help for any dataset statistically significantly. \textit{Separate} and \textit{Mentions} models are significantly better for \textit{Spanish} (\textit{Mentions} even for \textit{Catalan}) which has almost no singletons. One reasonable explanation is that the baseline model is under-parametrized for this dataset. We evaluated this hypothesis by increasing the model size, but the improvement was not statistically significant. In the following chapter, we evaluate singletons on the model with joined pretraining; we compute confidence intervals, and in Table \ref{tab:final_results}, we can see that the improvement is below statistical significance (on $95\%$ significance level). So, we believe that these strange results in Table \ref{tab:singletons_results} are caused by random chance.

Another important conclusion from this table is that for all the datasets that have some singletons, the singletons model achieves better results than the baseline.

The reason is probably in a better mention identification due to more training data. We also analyzed an additional metric, mention overlap ratio (MOR), which measures token overlap between gold and system mentions. Singleton modeling increased MOR from 65\% to 73\% compared to baseline. We investigated concrete examples to validate this hypothesis. An example from the English GUM dataset follows:
\begin{description}
    \item[baseline] \textit{Research on \textcolor{red}{adult-learned second language (L2$_{e1}$}) has provided considerable insight into the neurocognitive mechanisms underlying the learning and processing of \textcolor{red}{L2$_{e1}$} \textcolor{blue}{grammar$_{e2}$}.}
    \item[singletons/GOLD] \textit{\textcolor{red}{Research on} \textcolor{blue}{adult-learned second language$_{e2}$} \textcolor{red}{(}\textcolor{blue}{L2$_{e2}$}\textcolor{red}{)$_{e1}$} has provided \textcolor{green}{considerable insight into} \textcolor{cyan}{the neurocognitive mechanisms underlying} \textcolor{magenta}{the learning and processing of} \textcolor{blue}{L2$_{e2}$} \textcolor{orange}{grammar$_{e6}$}\textcolor{magenta}{$_{e5}$}\textcolor{cyan}{$_{e4}$}\textcolor{green}{$_{e3}$}.} 
\end{description}

We can see that the model with singletons can handle even complicated nested mentions, whereas the baseline cannot.

\subsubsection{Headword Representation}

We went through some examples on the datasets where heads-only representation helps very significantly. For example, in Czech PDT, we found that there are many very long entity mentions that are not hard to capture with the standard span-based model. Sometimes a mention does not even fit into the maximum span length. We believe this is the main reason why the head-based model performs so much better. For example, in the sentence from the Czech PDT corpus: 

\textit{"Ústavní soud rozhodl ve středu o podnětu skupiny 19 členů pražských zastupitelstev na vypuštění té části zákona o volbách do obecních zastupitelstev, která podmiňuje registraci sdružení nezávislých kandidátů předložením petice podepsané sedmi procenty obyvatel obce."}

In English:

\textit{"On Wednesday the Constitutional Court ruled on a petition filed by a group of 19 members of the Prague City Council to strike out the part of the law on municipal elections that makes the registration of an association of independent candidates contingent on the submission of a petition signed by seven percent of the municipality's residents."}

The whole sentence is a single entity representing the concrete decision of the court, and there are many mentions of this decision later in the document.

\subsubsection{Best Combinations}
\label{sec:discussion_best}

\begin{table*}
    \centering

\begin{adjustbox}{width=\linewidth}
\begin{tabular}{@{}lrrrrr|r@{}}
\toprule
Dataset/Extension & \multicolumn{1}{l}{Joined} & \multicolumn{1}{l}{Trees}                       & \multicolumn{1}{c}{Span2head}                    & \multicolumn{1}{l}{Singletons}                   & \multicolumn{1}{l|}{Only Heads}                                        & \multicolumn{1}{l}{Best Comb.} \\ \midrule
ca\_ancora        & 74.42 $\pm$ .25            & 74.17 $\pm$ .27                                 & 72.65 $\pm$ .39                                  & {\ul 74.93 $\pm$ .33}                            & \cellcolor[HTML]{009901}{\textbf{82.22 $\pm$ .4}} & 82.12 $\pm$ .13                      \\
cs\_pcedt         & 73.23 $\pm$ .08            & 73.2 $\pm$ .1                                   & 71.27 $\pm$ .11                                  & 73.17 $\pm$ .22                                  & \cellcolor[HTML]{009901}\textbf{75.95 $\pm$ .13}                       & 75.89 $\pm$ .09                      \\
cs\_pdt           & 73.47 $\pm$ .23            & 73.2 $\pm$ .46                                  & 72.98 $\pm$ .29                                  & 73.37 $\pm$ .28                                  & \cellcolor[HTML]{009901}\textbf{79.43 $\pm$ .1}                        & 79.51 $\pm$ .02                      \\
de\_parcorfull    & 76.57 $\pm$ 1.5            & \cellcolor[HTML]{009901}\textbf{79.4 $\pm$ 1.1} & 76.35 $\pm$ 1.8                                  & \cellcolor[HTML]{009901}\textbf{79.57 $\pm$ .93} & \cellcolor[HTML]{009901}{\ul 78.63 $\pm$ 1.9}                          & 79.13 $\pm$ .89                      \\
de\_potsdamcc     & 76.3 $\pm$ 1.1             & \cellcolor[HTML]{009901}{\ul 78.5 $\pm$ 1.6}    & 76.28 $\pm$ .94                                  & 76.62 $\pm$ 1.2                                  & 75.16 $\pm$ .79                                                        & 78.41 $\pm$ .89                      \\
en\_gum           & 69.04 $\pm$ .49            & \cellcolor[HTML]{009901}{\ul 70.19 $\pm$ .75}   & \cellcolor[HTML]{FFFFFF}\textbf{70.84 $\pm$ .27} & \cellcolor[HTML]{009901}\textbf{71.29 $\pm$ .62} & \cellcolor[HTML]{009901}\textbf{75.18 $\pm$ .11}                       & \textbf{75.51 $\pm$ .16}             \\
en\_parcorfull    & 69.16 $\pm$ 2.0            & \cellcolor[HTML]{009901}\textbf{74.1 $\pm$ 2.7} & 67.19 $\pm$ 1.4                                  & {\ul 70.4 $\pm$ 2.3}                             & 61.77 $\pm$ 1.4                                                        & 72.8 $\pm$ .2.5                      \\
es\_ancora        & 76.09 $\pm$ .13            & 76.07 $\pm$ .15                                 & 72.97 $\pm$ .28                                  & {\ul 76.2 $\pm$ .18}                             & \cellcolor[HTML]{009901}\textbf{82.39 $\pm$ .08}                       & 82.43 $\pm$ .07                      \\
fr\_democrat      & 65.24 $\pm$ .49            & 65.46 $\pm$ .4                                  & 65.38 $\pm$ .49                                  & \cellcolor[HTML]{009901}\textbf{66.23 $\pm$ .44} & \cellcolor[HTML]{009901}\textbf{67.72 $\pm$ .81}                       & \textbf{68.58 $\pm$ .23}             \\
hu\_korkor        & 68.99 $\pm$ .95            & 67.75 $\pm$ 1.5                                 & \textbf{71 $\pm$ .69 }                              & {\ul 69.82 $\pm$ 1.5}                            & \cellcolor[HTML]{009901}\textbf{73.51 $\pm$ .86}                       & 73.55 $\pm$ .57                      \\
hu\_szegedkoref   & 69.52 $\pm$ .59            & 69.59 $\pm$ .54                                 & 69.32 $\pm$ .34                                  & 69.47 $\pm$ .82                                  & \cellcolor[HTML]{009901}\textbf{70.68 $\pm$ .46}                       & 70.67 $\pm$ .31                      \\
lt\_lcc           & 73.59 $\pm$ 1.1            & {\ul 74.7 $\pm$ .72}                            & 73.65 $\pm$ .62                                  & \cellcolor[HTML]{009901}\textbf{76.28 $\pm$ .79} & \cellcolor[HTML]{009901}\textbf{76.77 $\pm$ .98}                       & {\ul 77.65 $\pm$ .71}                \\
no\_bokmaalnarc   & 74.79 $\pm$ .44            & {\ul 75.48 $\pm$ .49}                           & {\ul 75.64 $\pm$ .66}                            & 74.67 $\pm$ .24                                  & \cellcolor[HTML]{009901}\textbf{77.84 $\pm$ .29}                       & 78.15 $\pm$ .29                      \\
no\_nynorsknarc   & 75.98 $\pm$ .17            & 74.79 $\pm$ .48                                 & 74.96 $\pm$ 1                                    & \cellcolor[HTML]{009901}\textbf{76.59 $\pm$ .42} & \cellcolor[HTML]{009901}\textbf{78.51 $\pm$ .21}                       & \textbf{78.81 $\pm$ .16}             \\
pl\_pcc           & 74.1 $\pm$ .23             & {\ul 74.35 $\pm$ .13}                           & 74.19 $\pm$ .24                                  & \cellcolor[HTML]{009901}\textbf{75.22 $\pm$ .2}  & \cellcolor[HTML]{009901}\textbf{76.1 $\pm$ .11}                        & 76 $\pm$ .17                         \\
ru\_rucor         & 70.38 $\pm$ .64            & 70.23 $\pm$ .42                                 & 70.49 $\pm$ .42                                  & 70.6 $\pm$ .41                                   & \cellcolor[HTML]{009901}\textbf{75.8 $\pm$ .58}                        & 75.98 $\pm$ .42                      \\
tr\_itcc          & 45.39 $\pm$ 1.3            & 34.79 $\pm$ 7.7                                 & 29.53 $\pm$ .5.2                                 & 45.63 $\pm$ .4                                   & 45.17 $\pm$ 3.9                                                        & 44.17 $\pm$ 1.8                      \\ \midrule
avg               & 70.95                      & 70.94                                           & 69.69                                            & 71.77                                            & 73.70 & 74.67                 \\ \bottomrule
\end{tabular}
\end{adjustbox}
\caption{Performance gains over the best base model with proposed extensions together with 95\% confidence intervals. The models that surpass the baseline significantly are bold. The results, which are better only on lower confidence levels, are underlined. Variants selected for the best models are green.}
\label{tab:results_ext}
\end{table*}

Table \ref{tab:results_ext} presents the performance of all the extensions proposed. The table also shows the 95\% confidence intervals. \textit{Span2head} and \textit{Only head} extend \textit{Trees} (both also uses syntax tree representations). We can see, that tree representations alone improve the results only for \textit{ParCor} corpora. \textit{Span2Head} model helps significantly only for \textit{English-GUM} and \textit{Hungarian-korkor} corpora but it is not better than \textit{only heads} model. \textit{Singletons} modeling helps for most of the datasets that have singletons annotated. The exceptions are in the Norwegian Bokmaal and German Potsdam datasets. Another interesting point is that for the German ParCor dataset, singletons modeling helps significantly even though de-parcor itself does not contain singletons. One possible explanation is in the joined model. Adding the singletons from other languages to the training data can help to predict the mentions in the German ParCor dataset. Results in Table \ref{tab:singletons_results} support this theory because we can see that when we train the model from scratch, singletons modeling does not help. \textit{Only heads} model helps for almost all the datasets since it reduces the mention space and improves precision significantly. The extensions selected for the best model are highlighted in green color and the last column shows the results of the best combination. Please note that the results may show slight inconsistencies, even for datasets where a single extension was selected (e.g., 'Only heads' and 'Best Comb.' for Hungarian). This is because the experiments were rerun, averaging results across different runs with the same configuration. However, these differences are within the confidence intervals, ensuring the results remain reliable.

\subsubsection{Overlapping Segments}

\begin{table*}
    \centering

\begin{adjustbox}{width=.85\linewidth}
\begin{tabular}{@{}lrrrrr@{}}
\toprule
max. segments 8                                     & None                  & min                      & min\_filter           & max                   & max\_filter              \\ \midrule
\multicolumn{1}{l|}{ca\_ancora}      & 82.39 $\pm$ .25       & {\ul 82.57 $\pm$ .25}    & {\ul 82.54 $\pm$ .24} & {\ul 82.5 $\pm$ .24}  & {\ul 82.5 $\pm$ .25}     \\
\multicolumn{1}{l|}{cs\_pcedt}       & 78.23 $\pm$ .13       & 78.18 $\pm$ .11          & 78.23 $\pm$ .15       & 78.19 $\pm$ .15       & 78.12 $\pm$ .1           \\
\multicolumn{1}{l|}{cs\_pdt}         & 80.02 $\pm$ .21       & 79.95 $\pm$ .19          & 79.97 $\pm$ .2        & 79.98 $\pm$ .2        & 79.96 $\pm$ .19          \\
\multicolumn{1}{l|}{de\_parcorfull}  & 81.54 $\pm$ 1.3       & 81.54 $\pm$ 1.3          & 81.54 $\pm$ 1.3       & 81.54 $\pm$ 1.3       & 81.54 $\pm$ 1.3          \\
\multicolumn{1}{l|}{de\_potsdamcc}   & 75.88 $\pm$ 2.2       & 75.88 $\pm$ 2.2          & 75.88 $\pm$ 2.2       & 75.88 $\pm$ 2.2       & 75.88 $\pm$ 2.2          \\
\multicolumn{1}{l|}{en\_gum}         & 76.25 $\pm$ .12       & 76.25 $\pm$ .12          & 76.25 $\pm$ .12       & 76.25 $\pm$ .12       & 76.25 $\pm$ .12          \\
\multicolumn{1}{l|}{en\_parcorfull}  & 67.92 $\pm$ 3.1       & 67.92 $\pm$ 3.1          & 67.92 $\pm$ 3.1       & 67.92 $\pm$ 3.1       & 67.92 $\pm$ 3.1          \\
\multicolumn{1}{l|}{es\_ancora}      & 82.65 $\pm$ .27       & 82.65 $\pm$ .27          & 82.65 $\pm$ .27       & 82.65 $\pm$ .27       & 82.65 $\pm$ .27          \\
\multicolumn{1}{l|}{fr\_democrat}    & {\ul 69.49 $\pm$ .46} & 68.44 $\pm$ .69          & 69.01 $\pm$ .44       & 69.03 $\pm$ .58       & 68.9 $\pm$ .91           \\
\multicolumn{1}{l|}{fr\_split}       & 69.44 $\pm$ .42       & {\ul 69.97 $\pm$ .15}    & {\ul 69.98 $\pm$ .20} & 69.43 $\pm$ .50       & 69.66 $\pm$ .48          \\
\multicolumn{1}{l|}{hu\_korkor}      & 73.41 $\pm$ .43       & 73.41 $\pm$ .43          & 73.41 $\pm$ .43       & 73.41 $\pm$ .43       & 73.41 $\pm$ .43          \\
\multicolumn{1}{l|}{hu\_szegedkoref} & 71.2 $\pm$ .14        & 71.2 $\pm$ .14           & 71.2 $\pm$ .14        & 71.2 $\pm$ .14        & 71.2 $\pm$ .14           \\
\multicolumn{1}{l|}{lt\_lcc-corefud} & 76.82 $\pm$ .41       & 76.82 $\pm$ .41          & 76.82 $\pm$ .41       & 76.82 $\pm$ .41       & 76.82 $\pm$ .41          \\
\multicolumn{1}{l|}{no\_bokmaalnarc} & 78.96 $\pm$ .37       & 78.96 $\pm$ .37          & 78.96 $\pm$ .37       & 78.96 $\pm$ .37       & 78.96 $\pm$ .37          \\
\multicolumn{1}{l|}{no\_nynorsknarc} & 80.5 $\pm$ .14        & 80.5 $\pm$ .14           & 80.5 $\pm$ .14        & 80.5 $\pm$ .14        & 80.5 $\pm$ .14           \\
\multicolumn{1}{l|}{pl\_pcc}         & 75.97 $\pm$ .17       & {\ul 76.14 $\pm$ .23}    & 76.05 $\pm$ .22       & 76.06 $\pm$ .23       & {\ul 76.1 $\pm$ .15}     \\
\multicolumn{1}{l|}{ru\_rucor}       & 75.27 $\pm$ .42       & \textbf{77.23 $\pm$ .63} & 76.65 $\pm$ .39       & 76.77 $\pm$ .39       & \textbf{77.19 $\pm$ .57} \\
\multicolumn{1}{l|}{tr\_itcc}                             & 44.32 $\pm$ 2.2       & {\ul 45.22 $\pm$ 1.5}    & {\ul 45.26 $\pm$ 1.6} & {\ul 45.22 $\pm$ 1.4} & {\ul 45.32 $\pm$ .98}    \\ \midrule
avg                                  & 74.46                 & 74.60                    & 74.60                 & 74.57                 & 74.60 \\
\bottomrule
\end{tabular}
\end{adjustbox}
\caption{Results of the best model configurations with different approaches to merge segments during the prediction step. Maximum 8 segments.}
    \label{tab:segments_results}
\end{table*}

\begin{table*}
    \centering

\begin{adjustbox}{width=.85\linewidth}
\begin{tabular}{@{}l|rrrrr@{}}
\toprule
max. segments 4 & none             & min                      & min filter      & max              & max filter      \\ \midrule
ca\_ancora      & 82.39 $\pm$ .24  & {\ul 82.52 $\pm$ .25}    & 82.5 $\pm$ .22  & 82.42 $\pm$ .24  & 82.21 $\pm$ .24 \\
cs\_pcedt       & 77.82 $\pm$ .088 & 77.29 $\pm$ .17          & 77.47 $\pm$ .17 & 77.84 $\pm$ .096 & 77.65 $\pm$ .11 \\
cs\_pdt         & 79.9 $\pm$ .22   & 79.87 $\pm$ .16          & 79.84 $\pm$ .2  & 79.91 $\pm$ .22  & 79.74 $\pm$ .19 \\
de\_parcorfull  & 81.54 $\pm$ 1.3  & 81.54 $\pm$ 1.3          & 81.54 $\pm$ 1.3 & 81.54 $\pm$ 1.3  & 81.54 $\pm$ 1.3 \\
de\_potsdamcc   & 75.88 $\pm$ 2.2  & 75.88 $\pm$ 2.2          & 75.88 $\pm$ 2.2 & 75.88 $\pm$ 2.2  & 75.88 $\pm$ 2.2 \\
en\_gum         & 76.25 $\pm$ .12  & 76.25 $\pm$ .12          & 76.25 $\pm$ .12 & 76.25 $\pm$ .12  & 76.25 $\pm$ .12 \\
en\_parcorfull  & 67.92 $\pm$ 3.1  & 67.92 $\pm$ 3.1          & 67.92 $\pm$ 3.1 & 67.92 $\pm$ 3.1  & 67.92 $\pm$ 3.1 \\
es\_ancora      & 82.59 $\pm$ .27  & 82.48 $\pm$ .31          & 82.56 $\pm$ .26 & 82.59 $\pm$ .27  & 82.57 $\pm$ .28 \\
fr\_democrat    & 68.81 $\pm$ .84  & 68.64 $\pm$ .89          & 68.33 $\pm$ .79 & 68.81 $\pm$ .92  & 67.77 $\pm$ 1.7 \\
fr\_split       & 69 $\pm$ .6      & {\ul 69.46 $\pm$ .59}    & {\ul 69.48 $\pm$ .72} & 67.83 $\pm$ .72  & 68.75 $\pm$ .33 \\
hu\_korkor      & 73.41 $\pm$ .43  & 73.41 $\pm$ .43          & 73.41 $\pm$ .43 & 73.41 $\pm$ .43  & 73.41 $\pm$ .43 \\
hu\_szegedkoref & 71.2 $\pm$ .14   & 71.2 $\pm$ .14           & 71.2 $\pm$ .14  & 71.2 $\pm$ .14   & 71.2 $\pm$ .14  \\
lt\_lcc         & 76.82 $\pm$ .41  & 76.82 $\pm$ .41          & 76.82 $\pm$ .41 & 76.82 $\pm$ .41  & 76.82 $\pm$ .41 \\
no\_bokmaalnarc & 78.96 $\pm$ .36  & 78.91 $\pm$ .38          & 78.87 $\pm$ .34 & 78.96 $\pm$ .36  & 78.47 $\pm$ .37 \\
no\_nynorsknarc & 79.87 $\pm$ .57  & \textbf{80.32 $\pm$ .18} & 80.15 $\pm$ .37 & 79.87 $\pm$ .57  & 78.94 $\pm$ .12 \\
pl\_pcc         & 75.94 $\pm$ .15  & 76.05 $\pm$ .18          & 76.03 $\pm$ .22 & 75.93 $\pm$ .15  & 75.71 $\pm$ .17 \\
ru\_rucor       & 75.58 $\pm$ .37  & \textbf{76.54 $\pm$ .31} & 75.89 $\pm$ .55 & 75.62 $\pm$ .42  & 72.8 $\pm$ .25  \\
tr\_itcc        & 43.61 $\pm$ 1.5  & 43.85 $\pm$ 1.8          & 43.81 $\pm$ 1.3 & 43.59 $\pm$ 1.5  & 42.19 $\pm$ 2.7 \\ \midrule
avg                                  & 74.31            & 74.39                    & 74.33                 & 74.24            & 73.88 \\ \bottomrule
\end{tabular}
\end{adjustbox}
\caption{Results of the best model configurations with different approaches to merge segments during the prediction step. Maximum 4 segments.}
    \label{tab:segments_results4}
\end{table*}

\begin{table*}
\centering
\begin{adjustbox}{width=.8\linewidth}
    
\begin{tabular}{@{}l|rrrrrr@{}}
\toprule
                & \multicolumn{2}{c|}{\begin{tabular}[c]{@{}c@{}}cross N segment corefs\\ {[}\%{]}\end{tabular}} & \multicolumn{2}{c|}{\begin{tabular}[c]{@{}c@{}}nearest coref cross N\\ {[}\%{]}\end{tabular}} & \multicolumn{2}{c}{\begin{tabular}[c]{@{}c@{}}segments over N\\ {[}\%{]}\end{tabular}} \\ \cmidrule(l){2-7} 
                & \multicolumn{1}{c}{4}                          & \multicolumn{1}{c|}{8}                        & \multicolumn{1}{c}{4}                         & \multicolumn{1}{c|}{8}                        & \multicolumn{1}{c}{4}                     & \multicolumn{1}{c}{8}                     \\ \midrule
ca\_ancora      & 5.975                                          & 1.459                                         & .6044                                         & .2747                                         & 4.739                                     & 1.422                                     \\
cs\_pcedt       & 15.82                                          & 7.398                                         & 1.107                                         & .1441                                         & 13.55                                     & 2.133                                     \\
cs\_pdt         & 9.787                                          & 1.54                                          & .4787                                         & .06769                                        & 4.175                                     & .3976                                     \\
de\_parcorfull  & 0.0                    & 0.0                   & 0.0                   & 0.0                   & 0.0               & 0.0               \\
de\_potsdamcc   & 0.0                    & 0.0                   & 0.0                   & 0.0                   & 0.0               & 0.0               \\
en\_gum         & 0.0                    & 0.0                   & 0.0                   & 0.0                   & 0.0               & 0.0               \\
en\_parcorfull  & 0.0                    & 0.0                   & 0.0                   & 0.0                   & 0.0               & 0.0               \\
es\_ancora      & 3.012                                          & 0.0                                           & .1396                                         & 0.0                                           & .5076                                     & 0.0                                       \\
fr\_democrat    & 52.86                                          & 25.12                                         & 1.664                                         & .8106                                         & 34.86/27.52                                     & 27.52/12.84                                     \\
hu\_korkor      & 0.0                    & 0.0                   & 0.0                   & 0.0                   & 0.0               & 0.0               \\
hu\_szegedkoref & 0.0                    & 0.0                   & 0.0                   & 0.0                   & 0.0               & 0.0               \\
lt\_lcc         & 0.0                    & 0.0                   & 0.0                   & 0.0                   & 0.0               & 0.0               \\
no\_bokmaalnarc & 9.026                                          & 0.0                                           & .1092                                         & 0.0                                           & 3.846                                     & 0.0                                       \\
no\_nynorsknarc & 15.49                                          & 0.0                                           & .5405                                         & 0.0                                           & 10.29                                     & 0.0                                       \\
pl\_pcc         & 11.79                                          & 2.724                                         & .3591                                         & .1554                                         & 2.954                                     & .8439                                     \\
ru\_rucor       & 43.77                                          & 21.76                                         & 3.273                                         & 1.283                                         & 35.44                                     & 16.46                                     \\
tr\_itcc        & 55.6                                           & 25.07                                         & 5.703                                         & 2.037                                         & 55.56                                     & 11.11                                     \\ \bottomrule
\end{tabular}
\end{adjustbox}
\caption{Statistics of long documents and distant coreference relations across the datasets.}
\label{tab:long_stats}
\end{table*}

Table \ref{tab:segments_results} evaluates longer context implementation through overlapping segments. We use a maximum length of 8 segments\footnote{since we use this maximum length in most of our other experiments}, and we evaluate minimum and maximum overlap (1 and 7, respectively). We also try variants with filtering already-seen mentions as described in section \ref{sec:overlapping} (columns \textit{min\_filter} and \textit{max\_filter}). For most of the datasets, the results do not change at all, so additionally, we run the same experiment with just 4 segments (Table \ref{tab:segments_results4})\footnote{which is the minimum number of segments we used in our models}, and we count basic statistics of long coreference relations (Table \ref{tab:long_stats}). Table \ref{tab:long_stats} presents three types of information: \begin{enumerate}
    \item Column \textit{cross N segment corefs} shows the percentage of coreference links that cross the boundary of N segments. These links cannot be predicted without the overlapping segments.
    \item Column \textit{nearest coref cross N} is the percentage of mentions for which the nearest antecedent is already in a different segments block, so all the antecedents are in the different block, and the antecedent cannot be predicted with the standard model.
    \item Column \textit{segments over N} shows the percentage of segments that are after the first N-segments block. Only those segments beyond the first N-segment block are affected by predictions involving segment overlapping, as the remaining segments fit within the model's capacity and are processed in a single example.
\end{enumerate}

A significant number of log-distant coreferences are in French, Turkish, and Russian datasets. Some of them are also in both Czech datasets, Catalan and Polish. By far the most long coreferences are in the French dataset. From Table \ref{tab:segments_results}, we can see that segments overlapping decreases the performance. We investigated the dataset closely and found out that in this dataset, multiple documents are merged into a single document quite often. When this happens, cross coreferences between those concatenated documents are not annotated so the model which can predict long-distant coreferences is penalized. To evaluate the models correctly, we add the variant where we split the documents in French dev data manually (row \textit{fr\_split}). After splitting, results with segments overlapping improve significantly, but they are still not as high as we expected in comparison with Russian, where the overlap helps significantly. To evaluate this a bit further, we also computed the recall for cross-segment coreference links. The recall is 72\% for Russian and only 44\% for French. By comparing these datasets manually, we conclude that the coreference chains are very long in the French dataset, and it is probably too hard to predict them correctly from the antecedents (one mistake in the chain might have a huge effect on the overall performance). For the Turkish dataset, it is hard to make any significant improvement due to its low quality.

\subsection{Cross-lingual Transfer}

\begin{table*}
    \centering
\scalebox{1}{
\begin{tabular}{@{}lrlll@{}}
\toprule
Dataset Zero-shot & \multicolumn{1}{l}{Finetunned} & Joined               & \multicolumn{1}{c}{Zero-shot} & Lang. Zero-shot       \\ \midrule
ca\_ancora        & 82.12 $\pm$ .13                & 81.9225 $\pm$ 0.38   & 72.01 $\pm$ .17               & 58.53 $\pm$ .49       \\
cs\_pcedt         & \textbf{75.89 $\pm$ .09}                & 74.7775 $\pm$ 0.23   & 62.26 $\pm$ .02               & 51.78 $\pm$ .24       \\
cs\_pdt           & \textbf{79.51 $\pm$ .02}                & 78.8575 $\pm$ 0.25   & 69.92 $\pm$ .16               & 64.55 $\pm$ .28       \\
de\_parcorfull    & \textbf{79.13 $\pm$ .89}                & 65.095 $\pm$ 3.02    & 56.77 $\pm$ .82               & {\ul 62.33 $\pm$ .3}  \\
de\_potsdamcc     & \textbf{78.41 $\pm$ .89}                & 75.1325 $\pm$ 2.24   & 67.15 $\pm$ 1.4               & 64.33 $\pm$ .64       \\
en\_gum           & \textbf{75.51 $\pm$ .16}                & 74.71 $\pm$ 0.41     & 60.02 $\pm$ .39               & {\ul 62.03 $\pm$ .38} \\
en\_parcorfull    & \textbf{72.8 $\pm$ .2.5}                & 46.3575 $\pm$ 1.29   & {\ul 58.06 $\pm$ 1.9 }              & 46.67 $\pm$ .81       \\
es\_ancora        & 82.43 $\pm$ .07                & 82.245 $\pm$ 0.35    & 75.09 $\pm$ .26               & 61.68 $\pm$ .38       \\
fr\_democrat      & \textbf{68.58 $\pm$ .23}                & 67.7225 $\pm$ 0.5    & 60.32 $\pm$ .14               & -                     \\
hu\_korkor        & 73.55 $\pm$ .57                & 73.2625 $\pm$ 0.54   & 58.26 $\pm$ .31               & 53.75 $\pm$ .35       \\
hu\_szegedkoref   & 70.67 $\pm$ .31                & 69.75 $\pm$ 0.65     & 55.16 $\pm$ .11               & 53.36 $\pm$ .17       \\
lt\_lcc           & \textbf{77.65 $\pm$ .71}                & 75.65 $\pm$ 0.78     & 47.93 $\pm$ .59               & -                     \\
no\_bokmaalnarc   & 78.15 $\pm$ .29                & 77.8225 $\pm$ 0.67   & 74.83 $\pm$ .05               & 64.77 $\pm$ .76       \\
no\_nynorsknarc   & \textbf{78.81 $\pm$ .16}                & 77.955 $\pm$ 0.46    & 72.7 $\pm$ .37                & 63.24 $\pm$ .94       \\
pl\_pcc           & 76 $\pm$ .17                   & 76.0625 $\pm$ 0.08   & 58.44 $\pm$ .13               & -                     \\
ru\_rucor         & \textbf{75.98 $\pm$ .42}                & 72.665 $\pm$ 0.83    & 62.12 $\pm$ .28               & -                     \\
tr\_itcc          & 44.17 $\pm$ 1.8                & 38.52 $\pm$ 1.61     & \textbf{46.26 $\pm$ .2}       & -                     \\ \midrule
avg               & \multicolumn{1}{l}{74.67}      & \multicolumn{1}{l}{71.09} & 62.19                         & 58.92                 \\ \bottomrule
\end{tabular}
}
\caption{Zero-shot cross-lingual transfer results. The table compares four settings. A joined model trained for all datasets, a joined model fine-tuned on each dataset separately afterward, a zero-shot setting for each dataset and zero-shot setting for each language.}
    \label{tab:zero_results}
\end{table*}

Table \ref{tab:zero_results} shows the results of cross-lingual transfer evaluation. We can see that for most of the datasets, fine-tuning the model on the specific data helps, but the difference is not so significant. For several datasets, the results are even comparable. For small datasets, the fine-tuned models achieve much better results than the joined model. Note that we do not use any weighing of particular datasets in the joined collection, so small datasets have a marginal effect on the training. The results of dataset zero-shot and language zero-shot are mostly very predictable. For the languages with more datasets in CorefUD collection, the drop of zero-shot compared to the joined model was smaller than for the rest of the datasets. When we discard all the datasets for each language, the results are mostly significantly worse compared to discarding just a single dataset. This behavior is expected. Several exceptions should be analyzed more. The strangest results are on the Turkish dataset. Here, the zero-shot model achieves the best performance. Together with the fact that the numbers are very low for this dataset, it suggests that there is some noise in the dataset annotation.  If we train the model on the training part of the Turkish dataset the noise is present on two sides (training and testing). If we train a zero-shot model, we eliminate one part. 

Another interesting aspect is the superior performance of \textit{de-parcor} and \textit{english-gum} datasets in the language zero-shot scenario.

\subsection{Final Results}

\begin{table*}
    \centering
\scalebox{1}{
\begin{tabular}{@{}lllll@{}}
\toprule
                & Our Best             & CorPipe        & CorPipe-large & wl-coref\\ \midrule
ca\_ancora      & \textbf{82.05}       & 79.93          & 82.39      & -   \\
cs\_pcedt       & {\ul \textbf{78.85}} & 76.02          & 77.93      & -   \\
cs\_pdt         & {\ul \textbf{78.23}} & 76.76          & 77.85      & -   \\
de\_parcorfull  & {\ul \textbf{77.62}} & 63.3           & 69.94      & -   \\
de\_potsdamcc   & 71.07                & \textbf{72.63} & 67.93      & -   \\
en\_gum         & \textbf{73.19}       & 72.33          & 75.02      & -   \\
en\_parcorfull  & \textbf{61.62}       & 57.58          & 64.79      & -   \\
es\_ancora      & {\ul \textbf{82.71}} & 81.18          & 82.26      & -   \\
fr\_democrat    & {\ul \textbf{69.59}} & 65.42          & 68.22      & -   \\
hu\_korkor      & {\ul \textbf{69.03}} & 66.18          & 67.95      & -   \\
hu\_szegedkoref & \textbf{69.02}       & 65.4           & 69.16      & -   \\
lt\_lcc         & \textbf{70.69}       & 68.63          & 75.63      & -   \\
no\_bokmaalnarc & \textbf{76.24}       & 75.43          & 78.94      & -   \\
no\_nynorsknarc & \textbf{74.54}       & 73.64          & 77.24      & -   \\
pl\_pcc         & 77                   & {\ul \textbf{79.04}} & 78.93   & -      \\
ru\_rucor       & {\ul \textbf{81.48}} & 78.43          & 80.35        & - \\
tr\_itcc        & {\ul \textbf{56.45}}        & 42.47          & 49.97 & -        \\
\midrule
avg             & {\ul \textbf{73.49}}                & 70.26          & 73.21 & -
\\
\midrule
OntoNotes & 80.5 & \textbf{80.7} & 82 & 81\\
\bottomrule
\end{tabular}
}
\caption{Final results on test sets. The proposed model compared to two Corpipe variants (without ensembling) CorPipe uses the same encoder as the proposed model (XLMR-large), and CorPipe-large uses a larger T5 model.}
    \label{tab:final_results}
\end{table*}

Table \ref{tab:final_results} presents the final results of the best-performing models on a test set. For comparison, we use the best model of CRAC 2023 Multilingual coreference resolution shared task from \cite{straka2023ufal} but not the best submission. The best model is much larger than ours, and it uses ensembling, so as the main model for comparison, we use their RemBERT version without ensembling, which has a comparable size to the proposed model. 

We can see that the proposed model outperformed CorPipe on most of the datasets. For some datasets (mostly the ones without singletons annotated), it outperformed even the large CorPipe model, which has approximately 3 times more trainable parameters. The only dataset where CorPipe outperformed our model by a large margin is the Polish one, which we consider an anomaly because, for this dataset, CorPipe results on the test set are much better than for the dev set. On the dev set, we achieve similar results. The same happened in the opposite way for the Russian dataset. Surprisingly, we even outperformed the large CorPipe model in average score, but this is caused mainly by a large margin on Turkish and German-parcor. 

We also evaluate our model on English OntoNotes dataset \cite{pradhan-etal-2012-conll} to validate that our model achieves reasonable performance. It should achieve similar results to word-level coreference since other improvements (like singletons modeling) are not important for OntoNotes. We can see that even though our model is slightly worse than CorPipe, the results are reasonable and close enough to the word-level coreference. We believe that the superior performance of wl-coref can be caused by using specific features for OntoNotes in wl-coref (like speaker information). Note that our goal was not to optimize the results on OntoNotes, so we did not optimize the model specifically for this corpus.


\section{Conclusion}

In this paper, we explored and evaluated various approaches to multilingual coreference resolution using the CorefUD 1.1 dataset. Our experiments revealed that monolingual models typically outperform cross-lingual models, especially for languages with datasets that are distinct in their characteristics. However, joint training across languages can still provide benefits for most of the datasets where sufficient cross-linguistic similarities exist.

We proposed several extensions to enhance the baseline models, including cross-lingual training, Span2Head modeling, syntactic information integration, headword mention representation, and long-context prediction. Among these, the heads-only model and singleton modeling showed the most consistent improvements across different datasets, demonstrating the importance of targeted adaptations for coreference resolution tasks. For several datasets, long-context prediction also brings a significant improvement. Additionally, our zero-shot cross-lingual experiments provided insights into the challenges and opportunities of cross-lingual transfer, with the Turkish dataset results highlighting potential issues related to noise in the training data.

Overall, our findings emphasize the need for tailored approaches in coreference resolution, particularly when dealing with diverse languages and annotation schemes. Future work could further investigate the impact of dataset-specific characteristics on model performance and explore additional strategies for enhancing cross-lingual transfer, especially for low-resource languages. Our source codes are publicly available for subsequent research\footnote{\url{https://github.com/ondfa/coref-multiling}}.

\paragraph{Acknowledgement}

Computational resources were supplied by the project ‘‘e-Infrastruktura
CZ’’ (e-INFRA LM2018140) provided within the program Projects of
Large Research, Development and Innovations Infrastructures.
This work has been supported by Grant No. SGS-2022-016 Advanced methods of data processing and analysis.

\bibliographystyle{unsrt}  
\bibliography{custom,anthology}

\end{document}